\documentclass[lettersize,journal]{IEEEtran}	
\usepackage{microtype}
\usepackage{graphicx}

\usepackage{subfigure}

\usepackage{hyperref}

\usepackage{amsmath}
\usepackage{amssymb}
\usepackage{mathtools}
\usepackage{verbatim}

\usepackage{mathrsfs}

\usepackage{algorithm}
\usepackage{algorithmic}

\usepackage{bm}

\usepackage{multirow}
\usepackage{enumerate}
\usepackage{amssymb}
\usepackage{makecell}
\usepackage{bbm}
\newcommand{\cO}{\mathcal{O}}
\newtheorem{theo}{\textbf{Theorem}}
\newtheorem{def1}{\textbf{Definition}}

\newtheorem{lem}[theo]{\textbf{Lemma}}

\newtheorem{cor}[theo]{\textbf{Corollary}}
\newtheorem{rem}{\textbf{Remark}}[section]

\newcommand*{\QEDA}{\hfill\ensuremath{\blacksquare}}%

\newcommand{\dataset}{{\cal D}}

\usepackage[textsize=tiny]{todonotes}

\begin{document}

\title{Linear RNNs Provably Learn Linear Dynamic Systems}

\author{Lifu Wang, Tianyu Wang*\thanks{Corresponding author: Tianyu Wang: TY\_Wang1992@foxmail.com}, Shengwei Yi, Bo Shen, Bo Hu, Xing Cao \thanks{Lifu Wang, Tianyu Wang and Shengwei Yi are with China Information Technology Security Evaluation Center (CNITSEC), Beijing, China.
Bo Shen, Bo Hu, Xing Cao are with Beijing Jiaotong University, Beijing, China.}}

% The paper headers
\markboth{Journal of \LaTeX\ Class Files,~Vol.~14, No.~8, August~2021}%
{Shell \MakeLowercase{\textit{et al.}}: A Sample Article Using IEEEtran.cls for IEEE Journals}

\maketitle
\begin{abstract}
We study the learning ability of linear recurrent neural networks with Gradient Descent. We prove the first theoretical guarantee on linear RNNs to learn any stable linear dynamic system using any a large type of loss functions. For an arbitrary stable linear system with a parameter $\rho_C$ related to the transition matrix $C$, we show that despite the non-convexity of the parameter optimization loss if the width of the RNN is large enough (and the required width in hidden layers does not rely on the length of the input sequence), a linear RNN can provably learn any stable linear dynamic system with the sample and time complexity polynomial in $\frac{1}{1-\rho_C}$.  Our results provide the first theoretical guarantee to learn a linear RNN and demonstrate how can the recurrent structure help to learn a dynamic system.
\end{abstract}

\section{Introduction}
Recurrent neural network(RNN) is a very important structure in machine learning to deal with sequence data. It is believed that using the recurrent structure, RNNs can lean complicated transformations of data over extended periods. Non-linear RNN has been proved to be Turing-Complete\cite{1995On}, thus can simulate arbitrary procedures. However, training RNN requires optimizing highly non-convex functions which are very hard to solve. On the other hand, it is widely believed\cite{2014Exact} that deep linear networks can capture some important aspects of optimization in deep learning and there are a series of recent papers trying to study the properties of deep linear networks \cite{2019Width,2020On,2016Deep}. Meanwhile, learning linear RNN itself is not only an important problem in System Identification but also useful for the language modeling in natural language processing \cite{BelangerK15}. In this paper, we study the non-convex optimization problem for learning linear RNNs.

Suppose there is a $d_p$-order and $d$-dimension  linear system  with  the following form:
\begin{equation}\label{st}
\begin{aligned}
&h_{t}=\bm{C}h_{t-1}+\bm{D}x_t,\\
&\widetilde{y}_t=\bm{G}h_t(x),\\
\end{aligned}
\end{equation}
where  $\bm{C}\in \mathbb{R}^{d_p\times d_p}, \bm{D}\in \mathbb{R}^{d_p\times d}$ and  $\bm{G}\in \mathbb{R}^{d_p\times d_y}$ are unknown system parameters.

At time $t$, this system output $\widetilde{y}_t$. It is nature to consider the system identification  problem to learn the unkonw system parameters from its outputs. We  consider a new linear (student) RNN with the form:
\begin{equation}\label{st2}
\begin{aligned}
&h'_{t}=\bm{W}h'_{t-1}+\bm{A}x_t,\\
&y'_t=\bm{B}h'_t(x),\\
\end{aligned}
\end{equation}
with  $\bm{W}\in \mathbb{R}^{m\times m}, \bm{A}\in \mathbb{R}^{m\times d}, \bm{B}\in \mathbb{R}^{m\times d_y}$ and train the parameters $\bm{W}, \bm{A}, \bm{B}$ to fit the output $\widetilde{y}_t$ of (\ref{st}).

Just like what is commonly done in machine learning, one may consider to collect data $\{x^i_t, \widetilde{y}^i_t\}$ then optimize the  empirical loss:
\begin{equation}
\begin{aligned}
L(\bm{W}, \bm{A}, \bm{B})&=\frac{1}{n\cdot T}\sum_i^n\sum_{t=1}^TL(\widetilde{y}^i_t,{y'_t}^i),\\
&=\frac{1}{n\cdot T}\sum_i^n\sum_{t=1}^TL(\bm{G}h_t(x^i), \bm{B}h'_t(x^i)),
\end{aligned}
\end{equation}
with Gradient Descent Algorithm, where $L$ is a convex loss function.

Therefore the following  questions arise naturally:
\begin{itemize}
\item \emph{Can gradient descent learn the target RNN in polynomial time and samples?}
\item  \emph{What kinds of random initializations(for example, how large widths will $\bm{W}$ be?)  do we need to learn the target RNN? }
 \end{itemize}

These problems look easy since it is very basic and important for the system identification problem. However, the loss $\frac{1}{n}\sum_i^n\frac{1}{T}\sum_{t=1}^TL(\widetilde{y}^i_t,{y'_t}^i)$ is {\bf non-convex} and   in fact even for SISO(single-input single-output, which means $x_t, y_t\in \mathbb{R}^1$) systems, this question is far from being trivial. Only after the work in \cite{2016Gradient}, the SISO case is solved. In fact, as pointed out in \cite{d6288499}, although the  widely used method in  system identification is the EM algorithm \cite{1999A}, yet it  is inherently non-convex and EM method will  get stuck in bad local minima easily.

One naive method is to only optimize the loss $L(\widetilde{y}_1,y'_1)$, which is a convex loss function. However, when $\widetilde{y}_t$ is not  accurately observed,  for example we can only observe $y_t=\widetilde{y}_t+n_t$ and $L(x,y)=||x-y||^2$ where $n_t$ is white noise with $\sigma$ variance at time $t$,   the results by optimizing $L(y_1,y'_1)$ may not be  optimal for the entire sequence loss $\frac{1}{T}\sum_{t=1}^TL(y_t, y'_t)$. In fact a naive estimate from $L(y_1, y'_1)$ will output an estimation with error $\sigma^2$ but $\frac{1}{T}\sum_{t=1}^TL(y_t, y'_t)$ may output an estimation with error $\cO(\sigma^2/\sqrt{T})$. It is shown in \cite{2016Gradient} that under some independent conditions of the inputs, SGD (stochastic gradient descent) converges to the global minimum of the maximum likelihood objective of an unknown linear time-invariant dynamical system from a sequence of noisy observations generated by the system and over-parameterization is helpful. However, their methods  heavily rely on the  SISO property ($x,y \in \mathbb{R}^1$) of the system, and  the condition $x_t$ for different $t$ are i.i.d. from Gaussian distribution. Their method  can not  be generalized to the systems with $x\in \mathbb{R}^d$ and $d>1$. It is still open  under which conditions can SGD  be guaranteed to find the global minimum of the linear RNN loss.

In this paper, we propose a new NTK method inspired by the work \cite{allenzhu2019sgd} and the authors' previous work\cite{WangSHC21} on non-linear RNN. And this is completely different method from that in \cite{2016Gradient} so we avoid the defect that the method in \cite{2016Gradient} can only be used in the SISO case.

We show that if the width $m$ of the linear RNN (\ref{st2}) is large enough (polynomial large), SGD can provably learn {\bf any} stable linear system {\bf with the sample and time complexity only polynomial in $\frac{1}{1-\rho_C}$ and independent of the input length $T$, where $\rho_C$ is roughly the spectral radius (see Section \ref{rh}) of the transition matrix $\bm{C}$}.  Learning linear RNN is a very important problem in System Identification. And since Gradient Descent with random initialization is the most commonly used method in machine learning, we are trying to understand this problem in a ``machine-learning style''. We believe  this can provide some insights for the recurrent structure in deep learning.

\section{Problem Formulation}
We consider the target linear system with the form:
\begin{equation}
\begin{aligned}
&p^i_{t}=\bm{C}p^i_{t-1}+\bm{D}x^i_t,\\
&\widetilde{y^i_t}=\bm{G}p^i_t,
\end{aligned}
\end{equation}
which is a stable  linear system with  $ ||\bm{C}^k\bm{D}||\leq c_\rho\cdot \rho_C^k$ for all $k\in \mathbb{N}$ and $\rho_C<1$, where $c_\rho>0$,  $||\bm{G}||, ||\bm{D}||= \Theta(1)$. For a given convex loss function $L$, we set the loss function
\begin{equation}
 L(\bm{C}, \bm{D}, \bm{G})=\mathbb{E}_{x,y \sim \dataset}\frac{1}{T}\sum_{t=1}^T L(y_t, \widetilde{y}_t).
\end{equation}
We define the global minimum $OPT_{\rho_C}$ as\\
$$OPT_{\rho_C}= \inf_{\bm{C}, \bm{D}, \bm{G}}\mathbb{E}_{x,y \sim \dataset}\frac{1}{T}\sum_{t=1}^T L(y_t, \widetilde{y}_t).$$
with $||\bm{C}^k\bm{D}||\leq c_\rho\rho_C^k,  k\in \mathbb{N}$, $c_0>0$ is an absolute constant.

Let  the  sequences $\{(x^i=(x^i_1,x^i_2,...,x^i_{T}))_{i=1}^K,\ (y^i=(y^i_1,y^i_2,...,y^i_{T}))_{i=1}^K\}$ be $K$ samples  i.i.d. drawn from $$\dataset =\{(x_1,x_2,...,x_{T})\times  (y_1,y_2,...,y_{T}) \in \mathbb{R}^{d\times T} \times \mathbb{R}^{d_y\times T}\}.$$  We consider a ``student'' linear system (RNN) to learn the target one. Let $f_t(\widetilde{\bm{W}}, \bm{A}, x^i)$ be the $t$-time output of a linear RNN with input $x^i$ and parameters $\widetilde{\bm{W}}\in \mathbb{R}^{m\times m}, \bm{A}\in \mathbb{R}^{m\times d}$:
\begin{equation}
\begin{aligned}
&h_{t}(x)=\widetilde{\bm{W}}h_{t-1}+\bm{A}x_t,\\
&\widetilde{f}_t(\widetilde{\bm{W}}, \bm{A}, x)=\bm{B}h_t(x).
\end{aligned}
\end{equation}
Our goal is to use $f_t(\widetilde{\bm{W}}, \bm{A}, x^i)$ and the $K$ samples to fit the empirical loss function and keeping the generalization error bound small.
\section{Our Result} The main result is formulated in the PAC-Learning setting as follow:
\begin{theo}\label{ti}(Informal) 
Under the conditions in the last section, suppose the entries of $\bm{W}$ in the student RNN are randomly initialized by i.i.d. generated from $N(0,\frac{\rho_C}{m})$. We use SGD algorithm to optimize. $\widetilde{\bm{W}}_{k}, \bm{A}_{k}$ are the $k$-th step outputs of SGD algorithm.

For any $\epsilon,\ \delta>0$,  and $ 0<\rho_C<1$, there exist parameters  $m^*=poly( \frac{1}{1-\rho_C},  \epsilon^{-1}, \delta^{-1},c_\rho)$ and $K=poly( \frac{1}{1-\rho_C},  \epsilon^{-1}, \delta^{-1},c_\rho)$ such that if $m>m^*$, with probability at least $1-\delta$,  SGD  can reach
\begin{equation}
\begin{aligned}
&\mathbb{E}_{x,y\sim \dataset }\frac{1}{K}\sum_{k=1}^K\frac{1}{T}\sum_{t=1}^T L(y_t, \widetilde{f}_t(\widetilde{\bm{W}}_k, \bm{A}_k, x))\\
&\leq OPT_{\rho_C} +\epsilon,
\end{aligned}
\end{equation}
 in $K$ steps.
\end{theo}
\begin{rem}
Theorem \ref{ti} induces  gradient descent with linear RNNs can provably learn any stable linear system with the iteration and sample complexity polynomial in $\frac{1}{1-\rho_C}$. This result is consistent with the previous Gradient Descent based method \cite{2016Gradient} to learn SISO linear systems.

In our theorem, all the parameters do not rely on the length $T$. Note that suppose at different time, $x^i_t$ are i.i.d. drawn from a distribution $\dataset'$, $n_t$ is the white noise and $T$ is large enough,
\begin{equation}\label{pred}
\begin{aligned}
&\mathbb{E}_{n_t, x_t\sim \dataset'}\frac{1}{T}\sum_{t=1}^T ||\widetilde{y}_t+n_t- \widetilde{f}_t(\widetilde{\bm{W}}, \bm{A}, x))||^2\\
&\leq \lim_{T'\to \infty}\mathbb{E}_{n_t, x_t\sim \dataset'}\frac{1}{T'}\sum_{t=1}^{T'} ||\widetilde{y}_t+n_t- \widetilde{f}_t(\widetilde{\bm{W}}, \bm{A}, x))||^2+\epsilon. \notag
\end{aligned}
\end{equation}
Thus optimizing the large  $T$ loss is enough to predict the complete dynamic behaviors of the target system.
\end{rem}
\subsection{Scale of $c_\rho$ and the  Comparison with Previous Results}\label{rh}
In our main theorem \ref{ti}, the parameters are polynomial in $c_\rho$. We should note that although we assume the target system is stable, this only means $$\rho(\bm{C})=\lim_{k\to\infty}||\bm{C}^k||^{1/k}<1.$$
In fact, suppose $\bm{C}\in \mathbb{R}^{N\times N}$, generally we have (see e.g. corollary 3.15 in \cite{power}):
$$||\bm{C}^k||\leq \sqrt{N}\sum_{j=0}^{N-1} \binom{N-1}{j}\binom{k}{j}\cdot||\bm{C}||_{\infty}^j\cdot \rho(\bm{C})^{k-j}.$$
When we set $\rho_C=\rho(\bm{C})$, $c_\rho$ can be very large and generally we should set $\rho_C>\rho(\bm{C})+\epsilon$ to make $c_\rho$ be polynomial in $N$.

On the other hand, the scale of $c_\rho$ is closely related to the so-called  ``acquiescent'' systems.
\begin{def1}
 Let $z\in \mathbb{C}$. A SISO $N$-order linear system with the transfer function  $\frac{s(z)}{p(z)}$ is called $\alpha$-acquiescent if 
 $$\{p(z)/z^N: |z|=\alpha \}\subseteq S,$$
 where $$S=\{z: Re(z)\geq (1+\tau_0)|\ Im(z)|\} \cap \{z: \tau_1\leq Re(z)\leq \tau_2\}$$
\end{def1}
And we have
\begin{lem}(Lemma 4.4  in \cite{2016Gradient}) Suppose the target linear system is SISO and $\alpha$-acquiescent. For any $k\in \mathbb{N}$,
\begin{equation}
    ||\bm{C}^k\bm{D}||\leq 2\pi n \alpha^{-2n}/\tau_1 \cdot \alpha^k.
\end{equation}
Thus in the SISO case, under the ``acquiescent'' conditions, our  Theorem \ref{ti} reduces to the main result in \cite{2016Gradient}.
\end{lem}
\begin{cor}(Corresponding to Theorem 5.1 in \cite{2016Gradient})
Supposing a SISO linear system is $\rho_C$-acquiescent, it is  learnable in polynomial time and polynomial samples.
\end{cor}
And for MIMO systems, our condition  $||\bm{C}^k\bm{D}||\leq c_\rho\rho_C^k,  k\in \mathbb{N}$ is a good generalization for MIMO systems.
%\begin{comment}

%\end{comment}

\subsection{Our Techniques} 
Our proof technique is closely related to the recent works on deep linear network \cite{2019Width}, non-linear network with neural tangent kernel \cite{DBLP}\cite{allenzhu2019convergence,cao}, and non-linear RNN\cite{allenzhu2019sgd}, \cite{rate}. Similar to \cite{DBLP}, we carefully upper and lower bound eigenvalues of this Gram matrix throughout the optimization process, using some perturbation analysis.  At the initialization point, we consider the spectral properties of Gaussian random matrices.  Using the linearization method, we can show these properties hold throughout the trajectory of gradient descent. Then we only need to construct a solution near the random initialization. And the distance from the solution to the initialization can be bounded by the stability of the system.

{\bf Notions.} For two matrices $\bm{A}, \bm{B}\in \mathbb{R}^{m\times n}$, we define $\langle \bm{A}, \bm{B} \rangle= \text{Tr}(A^TB)$. We define the asymptotic
notations $\cO(\cdot), \Omega(\cdot), \Theta(\cdot), poly(\cdot)$ as follows.  $a_n, b_n$ are two sequences. $a_n=\cO(b_n)$ if $\lim \sup_{n\to \infty}|a_n/b_n|< \infty$, $a_n=\Omega(b_n)$ if
$\lim \inf_{n\to \infty}|a_n/b_n|>0$, $a_n=\Theta(b_n)$ if $a_n=\Omega(b_n)$ and $a_n=\cO(b_n)$,  $a_n=poly(b_n)$ if there is $k\in \mathbb{N}$ that $a_n=O((b_n)^k)$. $\widetilde{\cO}(\cdot), \widetilde{\Omega}(\cdot), \widetilde{\Theta(\cdot)}, \widetilde{poly}(\cdot)$ are notions which hide the logarithmic factors in $\cO(\cdot), \Omega(\cdot), \Theta(\cdot) , poly(\cdot)$. $||\cdot ||$ and $||\cdot||_2$ denote the 2-norm of matrices. $||\cdot ||_1$ denotes the 1-norm. $||\cdot ||_F$ is the Frobenius-norm.
\section{Related Works}
{\bf Deep Linear Network.} The provable properties of the loss surface for deep linear networks were firstly shown in \cite{2016Deep}. In \cite{2019Width}, it is shown that if the width of the $L$-layer deep linear network is large enough (only depends on the output dimension, the rank $r$ and the condition number $\kappa$ of the input data), randomly initialized gradient descent will optimize deep linear networks in polynomial time in $L$, $r$ and  $\kappa$.
Moreover. in \cite{2020On}, the linear ResNet is studied and it is shown that Gradient Descent provably optimizes wide enough deep linear ResNets and the width does not rely on the number of layers.
%The techniques of these methods are closely related to the works on NTK  in  \cite{cao}, \cite{DBLP}, \cite{allenzhu2019sgd}, \cite{allenzhu2019convergence} and \cite{rate}
 %\begin{comment}

{\bf Over-Parameterization.}  Non-linear networks with one hidden node are studied in \cite{tian2017symmetry-breaking} and \cite{du2018when}. In these works, it is shown that, for a single-hidden-node ReLU network, under a very mild assumption on the input distribution, the loss is one point convex in a very large area. However, for the networks with multi-hidden nodes, the authors in \cite{safran2018spurious} pointed out that spurious local minima are common and indicated that an over-parameterization (the number of hidden nodes should be large) assumption is necessary.  Similarly, \cite{2016Gradient} showed that over-parameterization can help in the training process of a linear dynamic system. Another import progress is the theory about neural tangent kernel(NTK). The techniques of NTK for finite width network are studied in  \cite{cao}, \cite{DBLP}, \cite{allenzhu2019sgd}, \cite{allenzhu2019convergence} and \cite{rate}.
%\end{comment}

{\bf Learning Linear System.}
Prediction problems of time series for linear dynamical systems can be traced back to Kalman \cite{2009A}. In the case that the system is unknown, the first polynomial guarantees of running time and sample complexity bounds for learning single-input single-output (SISO) systems with gradient descent are provided in \cite{2016Gradient}. For MIMO systems. It was shown in \cite{165a59f}\cite{d6288499} that the spectral filtering method can be provably learned with polynomial guarantees of running time and sample complexity.\\

\section{Problem Setup and Main Results}
In this section, we introduce the basic problem setup and our main results.

Consider sequences $\{x^i=(x^i_1,x^i_2,...,x^i_{T})\}$ and the label $\{y^i=(y^i_1,y^i_2,...,y^i_{T})\}$ in the  data set. $x^i_t\in\mathbb{R}^d, y^i_t\in\mathbb{R}^{d_y}$ and $||x^i_t||, ||y^i_t||\leq \cO(1)$. We assume $d_y\leq d=\cO(1)$ and omit them in the asymptotic symbols.   We study  the linear RNN as: $\widetilde{\bm{W}}\in\mathbb{R}^{m\times m}, \bm{A}\in\mathbb{R}^{m\times d},\bm{B}\in\mathbb{R}^{d_y\times m}$
\begin{equation}\label{rnn}
\begin{aligned}
&h_{0}(x)=\bm{0}, h_{t}(x)=\widetilde{\bm{W}}h_{t-1}+\bm{A}x_t,\\
&\widetilde{f}_t(\widetilde{\bm{W}}, \bm{A}, x)=\bm{B}h_t(x)\in\mathbb{R}^{d_y}.
\end{aligned}
\end{equation}

 Assume $L^*(x)=L(y_t,x)$ is convex and locally Lipschitz convex function: for any $x,y$,  when $||x||\leq C, ||y_t||\leq C$,
\begin{equation}\label{lip}
||\nabla_x L^*(x)||\leq l_0(1+C).
\end{equation}
Then we perform algorithm \ref{a1}.
\begin{comment}
\begin{algorithm}
\caption{Learning Stable Linear System  with SGD}\label{a1}
{\bf Input:} Sequences of data $\{x,y\}$, learning rate $\eta$, initialization parameter $0<\rho<1$.\\
The entries of $\widetilde{\bm{W}}_0$ and $\bm{A}_0$ are i.i.d. generated from $N(0,\frac{\rho}{m})$ and $N(0,\frac{1}{m})$. The entries of $\bm{B}$ are i.i.d. generated from $N(0,\frac{1}{d_y})$. \\
\For {$k=0,1,2,3...K-1$}{
Randomly sample a sequence $x^i$ and the label $y^i$.

$\widetilde{\bm{W}}_{k+1}=\widetilde{\bm{W}}_{k}-\frac{\eta}{T}\sum_t\nabla_{\widetilde{\bm{W}_{k}}}L(y^i_{t},\widetilde{f}_{t}(\widetilde{\bm{W}_k}, \bm{A}_k, x^i))$,

$\bm{A}_{k+1}=\bm{A}_{k}-\frac{\eta}{T}\sum_t\nabla_{\bm{A}_{k}}L(y^i_{t},\widetilde{f}_{t}(\widetilde{\bm{W}}_k, \bm{A}_k, x^i))$.
}
\end{algorithm}
\end{comment}

\begin{algorithm}
  \caption{Learning Stable Linear System  with SGD}\label{a1}
\begin{algorithmic}
   \STATE {\bfseries Input:}Sequences of data $\{x,y\}$, learning rate $\eta$, initialization parameter $0<\rho<1$.
% \REPEAT
   \STATE {\bfseries Initialization:} The entries of $\widetilde{\bm{W}}_0$ and $\bm{A}_0$ are i.i.d. generated from $N(0,\frac{\rho}{m})$ and $N(0,\frac{1}{m})$. The entries of $\bm{B}$ are i.i.d. generated from $N(0,\frac{1}{d_y})$.
   \FOR{$k=0,1,2,3...K-1$}
   \STATE$ \widetilde{\bm{W}}_{k+1}=\widetilde{\bm{W}}_{k}-\frac{\eta}{T}\sum_t\nabla_{\widetilde{\bm{W}_{k}}}L(y^i_{t},\widetilde{f}_{t}(\widetilde{\bm{W}_k}, \bm{A}_k, x^i))$,
    \STATE    Randomly sample a sequence $x^i$ and the label $y^i$.
   \STATE $\bm{A}_{k+1}=\bm{A}_{k}-\frac{\eta}{T}\sum_t\nabla_{\bm{A}_{k}}L(y^i_{t},\widetilde{f}_{t}(\widetilde{\bm{W}}_k, \bm{A}_k, x^i))$.
   \ENDFOR
\end{algorithmic}
\end{algorithm}
\begin{comment}
On randomly generated data, we plot the curve of the square loss during training  for every mini-batch (batchsize=20) in Figure 1 with  $T=20$, $d=d_y=10$,  $\rho=0.9$, $\rho_C=0.8$. We can see SGD can easily find the global minima.
\begin{figure}
\centering
\includegraphics[width=0.4\textwidth]{f1.png}
\caption{Training Curve}\label{fig1}
\end{figure}
\end{comment}
In fact, we have
\begin{theo}\label{mt}
Assume there is $\delta\in [0,e^{-1}]$. Let $\rho_1=\frac{1}{1+10\cdot \frac{\log^2 m}{\sqrt{m}}}$, $0<\rho_0< 1$. Set the initialization parameter $\rho=\rho_1\cdot \rho^2_0$. Given an unknown distribution  $\dataset$  of sequences of $\{x, y\}$, let  $\widetilde{\bm{W}}_k$, $\bm{A}_k$ be the output of Algorithm \ref{a1}.

 For any small $\epsilon>0$, there are parameters \footnote{In order to simplify symbols, we omit $d$ and $d_y$ in these asymptotic bounds of parameters. One can easily show finally the parameters are  polynomial in $d$ and $d_y$. }
  \begin{equation}
  \begin{aligned}
 &T_{max}=\Theta(\frac{1}{\log(\frac{1}{\rho_0})} )\cdot \{ 2\log(\frac{c_\rho}{1-\rho_0})+\log(\frac{1}{\epsilon} )\\
&+\log \sqrt{\log(T_{max} /\delta)} +\frac{1}{2}\log(m) \}) \\
&b=\sqrt{ \log(T_{max} /\delta)}, \eta =\Theta(\frac{\nu \epsilon}{ mb^2}),\\
&K=\Theta(\frac{T^4_{max}b^4}{\nu \epsilon^2}),\\
 &m^*= \Theta(\frac{c_\rho^2 K^4  (1-\rho_0)^8 \epsilon^2  }{b^6})+\Theta(\frac{1}{\delta}),\\
 &\nu = \Theta(\frac{\epsilon^2(1-\rho_0)^{12}}{T_{max}^4\cdot  l_0^6(1+2b)^6}),
  \end{aligned}
  \end{equation}
  such that with probability at least $1-\delta$, if $m>m^*$,  for any $\bm{C}, \bm{D}, \bm{G}$, the algorithm outputs satisfy:
\begin{equation}
\begin{aligned}
\frac{1}{K}\sum_{k=0}^{K-1}&\frac{1}{T}\sum_{t=1}^T\{L(y^k_{t},\widetilde{f}_{t}(\widetilde{\bm{W}}_{k}, \bm{A}_{k}),x^k)\\
&- L(y^k_{t}, \widetilde{y}^k_{t}(\bm{C}, \bm{D}, \bm{G}))\}\leq \cO(\epsilon),
\end{aligned}
\end{equation}
where   $\widetilde{y}_t^k(\bm{C}, \bm{D}, \bm{G})$ is  the output from the linear system:
\begin{equation}
\begin{aligned}
&p^i_{t}=\bm{C}p^i_{t-1}+\bm{D}x^i_t,\\
&\widetilde{y^i_t}=\bm{G}p^i_t.
\end{aligned}
\end{equation}
with  $ ||\bm{C}^k\bm{D}||\leq c_\rho\rho_C^k$ for all $k\in \mathbb{N}$, $\bm{D}\in \mathbb{R}^{d_p\times d}$, $p_t\in \mathbb{R}^{d_p}$,  $\bm{C}\in \mathbb{R}^{d_p\times d_p} $ $\bm{G}\in \mathbb{R}^{d_y\times d_p}$,
\end{theo}
\begin{lem}
For the parameters of the last theorem, when $m>m^*$ and $\epsilon$ is small enough, we have the following results:
\begin{enumerate}
\item $\frac{T_{max}^4b^2}{Km\eta}= \Theta(\epsilon)$
\item $ l_0(1+2b)\cdot l_0^2(1+2b)^2\cdot \frac{\eta m \sqrt{K}}{(1-\rho_0)^6}\leq \cO(\epsilon)$
\item $K^2\eta^2\cdot \frac{m\sqrt{m}}{(1-\rho_0)^{11}})\cdot l_0^2(1+2b)^2\cdot l_0(1+2b)\leq \cO(\epsilon)$
\item $\frac{\eta m l^2_0(1+2b)^2}{(1-\rho_0)^6}\leq \cO(\epsilon)$
\item $\frac{ b\cdot d^2 c_\rho T^{3}_{max}\log m}{m^{1/2}}\leq\cO( \frac{\epsilon}{l_0(1+2b)})$
\end{enumerate}
\end{lem}
\begin{comment}
\begin{equation}
\begin{aligned}
&\eta m=\frac{\nu\cdot \epsilon}{b^2}\\
&r_0\cdot 16^2l_0^2(1+2b)^2\frac{\eta m \sqrt{K}}{(1-\rho_0)^6}\\
=&\nu\cdot \epsilon \cdot \sqrt{K} (1+\rho_0)^{-6}\cdot r_0 16^2l_0^2(1+2b)^2\\
=&\sqrt{\nu}\cdot T_{max} \sqrt{b}  (1+\rho_0)^{-6}\cdot r_0 16^2l_0^2(1+2b)^2\\
=&\epsilon
\end{aligned}
\end{equation}
\end{comment}

As for the population loss, we have:
\begin{theo}\label{gen}
(Rademacher complexity for RNN) Under the condition in Theorem \ref{mt}, with probability at least $1-\delta$,
\begin{equation}
\begin{aligned}
&|\mathbb{E}_{x,y\sim\dataset} \frac{1}{T}\sum_{t=1}^T\{L(y_{t},\widetilde{f}_{t}(\widetilde{\bm{W}}_{k}, \bm{A}_{k}, x))- L(y_{t}, \widetilde{y}_{t})\}\\
& -\frac{1}{K}\sum_{k=0}^{K-1}\frac{1}{T}\sum_{t=1}^T\{L(y^k_{t},\widetilde{f}_{t}(\widetilde{\bm{W}}_{k}, \bm{A}_{k}),x^k)- L(y^k_{t}, \widetilde{y}^k_{t})\}|\\
&\leq \cO(\epsilon). \notag
\end{aligned}
\end{equation}
\end{theo}
Since $\rho_1\to 1 $ as $m^*>poly(\frac{1}{1-\rho_0})$ large enough, we have $\rho_1>\rho_0$ and  $\rho > \rho_0^3$. Therefore from the above two theorems  we have the following corollary:
\begin{cor}
Let  the initialization parameter be $\rho_C<1$. For any small $\epsilon>0, \delta>0$, there is a parameter $m^*=poly(\frac{1}{\delta}, \frac{1}{\epsilon}, \frac{1}{1-\rho_C})$  such that with probability at least $1-\delta$, if $m>m^*, K=poly( \frac{1}{1-\rho_C}, \frac{1}{\epsilon}, \frac{1}{\delta})$,  the algorithm outputs satisfy:
\begin{equation}
\begin{aligned}
\mathbb{E}_{x,y\sim \dataset }&\frac{1}{T}\sum_{t=1}^T L(y_t, \widetilde{f}_t(\widetilde{\bm{W}}_k, \bm{A}_k, x))\leq OPT_{\rho_C} +\cO(\epsilon). \notag
\end{aligned}
\end{equation}
\end{cor}
\begin{rem}
In this paper, our results only assume for some constant $C$:
\begin{equation}
||\nabla_x L^*(x)||\leq \cO(1+C).
\end{equation}
This assumption is a very mild condition for the loss function $L(x)$, thus we, in fact, do not previously assume the form of the noise (for example, optimizing the square loss is to optimize the  maximum likelihood objective of the Gaussian noise, and $l^1$ loss is to optimize that for Laplace noise) and our result can even apply to not only the the regression problems but also the classification problems. In this aspect,  this result improves upon previous methods to learn linear dynamical systems in \cite{2016Gradient} and  \cite{d6288499} which highly rely on the form of the square loss.
\end{rem}
\section{Preliminary Properties}\label{pl}
Before proving Theorem \ref{mt} and \ref{gen}, we need some properties of Gaussian random matrices and linear RNNs. The proof is in the Supplementary Materials.

To simplify symbols, in the latter part of the paper, we set $\bm{W}_k=\widetilde{\bm{W}_k}/\rho$ and  \footnote{We omit $x$ in $f_t(\bm{W},\bm{A}, x)$ when it doesn't lead to misunderstanding.}
\begin{equation}
\begin{aligned}
f_t(\bm{W},\bm{A})=\widetilde{f}_t(\widetilde{\bm{W}},\bm{A}, x)=\sum _{t_0=0}^{t-1} \rho^{t_0}\bm{B} (\prod_{\tau=1}^{t_0} \bm{W})  \bm{A}X_{t-t_0}. \notag
\end{aligned}
\end{equation}
Then
\begin{equation}
\begin{aligned}
\bm{W}_{k+1}=&\bm{W}_{k}-\frac{\eta}{T\rho}\sum_{t=1}^T\nabla_{\widetilde{\bm{W}_{k}}}L(y^i_{t},\widetilde{f}_{t}(\widetilde{\bm{W}}_k, \bm{A}_k, x^i)),\\
=&\bm{W}_{k}-\frac{\eta}{T}\sum_{t=1}^T\nabla_{\bm{W}_{k}}L(y^i_{t},f_{t}(\bm{W}_k, \bm{A}_k, x^i)).
\end{aligned}
\end{equation}
Then we only need to consider $\bm{W}_k$. The entries of  $\bm{W}_0$ are i.i.d. generated from $N(0,\frac{1}{m})$. Let $B(\bm{W}_0,\omega)=\{\bm{W}|\ ||\bm{W}-\bm{W}_0||_F\leq \omega\}$ and $B(\bm{A}_0,\omega)=\{\bm{A}|\ ||\bm{A}-\bm{A}_0||_F\leq \omega\}$.
\subsection{Properties of Random Matrix}
One of the key points in this paper is that with high probability, the spectral radius of matrix $\bm{W}_0$ will be less than $\rho_1^{-1}$ and when $m\to \infty$, $\rho_1\to 1$. In fact, we have:
\begin{lem}\label{sp}
With probability at least $1-exp(-\Omega(\log^2m)))$ (it is larger than $ 1-\delta$ when $m>m^*$), there exists $L=c_0\cdot \frac{\sqrt{m}}{\log m}\in \mathbb{N}$ such that  for all $k \geq L$,  $$||\bm{W}_0^k||\leq \rho_1^{-k},$$ and for all $k<L$, $$||\bm{W}_0^k||\leq \rho_1^{-L},$$ where $c_0>1$ is an absolutely constant.
Meanwhile, for all $k\leq 2L$, with probability at least $1-exp(-\Omega(\log^2m)))$,
$$||W_0^k||\leq 2\sqrt{k}.$$
\end{lem}

In the rest of this paper, all the probabilities are considered under the condition  in Lemma \ref{sp}.

As a corollary, let $k\geq 2L$, $\omega_0= \frac{1}{\rho_0}-1$ and when $||\bm{W}-\bm{W}_0||_F\leq \omega\leq \omega_0 $, we have
\begin{equation}
\begin{aligned}
&||(\rho_1\cdot \rho_0^2 \cdot \bm{W})^k||= (\rho_1\cdot \rho_0^2)^k\sum_{i=0}^kC_{k}^i\cdot ||\bm{W}^i_0||\cdot ||\bm{W}-\bm{W}_0||^{k-i}\\
&= \sum_{i=0}^kC_{k}^i\cdot ||\bm{W}^i_0||\cdot (\rho_1\cdot \rho_0^2)^k\cdot (\frac{1}{\rho_0}-1)^{k-i}\\
&\leq \sum_{i=L}^kC_{k}^i(\rho_1\rho_0^2)^{i} [\rho_1\cdot(\rho_0-\rho_0^2)]^{k-i}\cdot \rho_1^{-k}\\
&+\sum_{i=0}^{L-1}C_{k}^i(\rho_1\cdot \rho_0^2)^{i}\cdot ||W_0^i||\cdot \rho_1^{k-i}\cdot (\rho_0-\rho_0^2)^{k-i} \leq \rho_0^k.
\end{aligned}
\end{equation}

For $k\leq 2L$,
\begin{equation}
\begin{aligned}
&||(\rho_1\cdot \rho_0^2 \cdot \bm{W})^k||= \rho_1^k\sum_{i=0}^kC_{k}^i\cdot ||\rho_0^2\bm{W}^i_0||\cdot ||\rho_0^2(\bm{W}-\bm{W}_0)||^{k-i}\\
&\leq \rho_1^k\sum_{i=0}^kC_{k}^i2\sqrt{i}\rho_0^{2i}\cdot (\rho_0-\rho_0^2)^{k-i}\\
&\leq \rho_1^k\rho_0^k2\sqrt{k}.
\end{aligned}
\end{equation}
Combing all these results, we can show the norm of $\rho^t \prod_{\tau=1}^{t}\bm{W}$ is bounded, which is:\\
For all $t\in \mathbb{N}$
\begin{equation}\label{wr}
||\rho^t \bm{W}^t||\leq 2\sqrt{t}\rho_0^{t}. 
\end{equation}

This is crucial to make the width independent on the length of input sequences. Then we can show:
\begin{lem}\label{lc} Let $\rho=\rho_1\cdot \rho_0^2$.
For any $\tau \in \mathbb{N}$, any $Z_t\in \mathbb{R}^d$ with $||Z_t|| =1$ and $\bm{Q}\in \mathbb{R}^{m\times d}, \bm{Q}_2\in \mathbb{R}^{m\times m}$, with probability at least $1-exp(-\Omega(\log^2m)))$, for any $\bm{W}\in B(\bm{W}_0,\omega)$, $\bm{A}\in B(\bm{A}_0,\omega)$ with $\omega\leq\omega_0$,
\begin{equation}
||\sum_{t=\tau}^{\infty} \rho^{t}\cdot \bm{B} (\prod_{\tau=1}^{t} \bm{W}) \bm{Q} Z_t|| \leq 4\frac{\sqrt{m}\tau(\rho_0)^{\tau}}{(1-\rho_0)^2}||\bm{Q}||,
\end{equation}
\begin{equation}
\begin{aligned}
&||\sum_{t_0=\tau}^\infty \sum_{t_1+t_2=t_0}  \rho^{t_0}\bm{B}(\prod_{\tau=1}^{t_1-1}\bm{W})  \bm{Q}_2  (\prod_{\tau=1}^{t_2-1} \bm{W})  \bm{A}Z_{t_0}||\\
&\leq 32\frac{\sqrt{m}\tau^2(\rho_0)^{\tau}}{(1-\rho_0)^3}||\bm{Q}_2||. \notag
\end{aligned}
\end{equation}
\end{lem}

Based on these results, we can only consider the properties of $\rho^{t_0}\bm{B} (\prod_{\tau'=1}^{t_0} \bm{W})  \bm{A}X_{t-t_0}$ for  bounded $t_0$. We have the following results:
\begin{lem}\label{lr}
For  any vector $v_1\in \mathbb{R}^{d_y}, v_2\in \mathbb{R}^d$ with $||v_1||=||v_2||=1$, if $m>\cO(\tau^3 \cdot d)$, with probability at least $1-exp(-\Omega(m/\tau^2))$:
\begin{equation}
\begin{aligned}
&0.9\leq ||\{\prod_{\tau'=1}^{t}\bm{W}_0\}\bm{A}_0v_2||\leq 1.1,\\
&0.9\leq ||\frac{1}{\sqrt{m}}\{\prod_{\tau'=1}^{t}\bm{W}_0\}^T\bm{B}^Tv_1||\leq 1.1,
\end{aligned}
\end{equation}
for any $0\leq t\leq \tau-1$ and $\tau>0$.
\end{lem}

\begin{lem}\label{lr2}
If $m>\Omega(\log(\tau \cdot d))$,  with probability at least $1-\delta$:
\begin{equation}
 ||\bm{B} (\prod_{\tau'=1}^{t} \bm{W})  \bm{A}||\leq  \sqrt{d\log(\tau\cdot d /\delta)}).
 \end{equation}
 for any $0\leq t\leq \tau-1$ and $\tau>0$.
\end{lem}

\begin{lem}\label{lr3}
For  any vector $v_1, u_1\in \mathbb{R}^{d_y}, v_2, u_2\in \mathbb{R}^d$  with $||v_1||=||u_1||=||v_2||=||u_2||=1$, if $m^{1/2}>\Omega(\tau^3 \cdot d)$, with probability at least $1-exp(-\Omega(\log^2 m))$, for any $0\leq t, t'\leq \tau-1$, $t\neq t'$:
\begin{equation}
\begin{aligned}
|(u_1)^T\bm{B}(\prod_{k=1}^{t}\bm{W}_0)&\cdot \{(\prod_{k=1}^{t'}\bm{W}_0)\}^T\bm{B}^Tv_1|\\
&\leq  24\tau d^2\log m,
\end{aligned}
\end{equation}
and
\begin{equation}
|(u_2)^T\bm{A}^T_0(\prod_{k=1}^{t}\bm{W}_0)^T\cdot \{(\prod_{k=1}^{t'}\bm{W}_0)\}\bm{A}_0v_2|\leq  24\tau\frac{d^2\log m} {\sqrt{m}}. \notag
\end{equation}
\end{lem}

\subsection{Properties of Linear RNN}
For any $t, \tau \in \mathbb{N}$,  $f_t$ can be written as (when $t-t_0<1$, we set $X_{t-t_0}=0$):
\begin{equation}
\begin{aligned}
f_t(\bm{W},\bm{A})=\sum _{t_0=0}^{t-1} \rho^{t_0}\bm{B} (\prod_{\tau'=1}^{t_0} \bm{W})  \bm{A}X_{t-t_0}.
\end{aligned}
\end{equation}
From Lemma \ref{lc}, we consider a truncation of $f_t$ as
\begin{equation}\label{tr}
\begin{aligned}
f^{\tau}_t(\bm{W},\bm{A})=\sum _{t_0=0}^{\tau} \rho^{t_0}\bm{B} (\prod_{\tau'=1}^{t_0} \bm{W})  \bm{A}X_{t-t_0}.
\end{aligned}
\end{equation}

$f_t(\bm{W},\bm{A})$ is also an almost linear function for $\bm{W}$ and $\bm{A}$. In fact we have:
\begin{lem}\label{l9}
Under the condition in Theorem \ref {mt},  with probability at least $1-\delta$, for all $t\in [T]$ and $\bm{W},  \bm{W}' \in B(\bm{W}_0, \omega)$, $\bm{A},  \bm{A}' \in B(\bm{A}_0, \omega)$ with $\omega \leq \omega_0$,
\begin{equation}
\begin{aligned}
&||f_t(\bm{W}',\bm{A}')-f_t(\bm{W},\bm{A})-\nabla_{W} f_t(\bm{W}, \bm{A})\cdot [\bm{W}'-\bm{W}]\\
& -\nabla_{A} f_t(\bm{W}, \bm{A})\cdot [\bm{A}'-\bm{A}] ||\\
&\leq  \Theta(\frac{\sqrt{m}\omega^2}{(1-\rho_0)^5}). \notag
\end{aligned}
\end{equation}
\end{lem}

Combing the above results, let
\begin{equation}
\begin{aligned}
f_t^{lin}(\bm{W}, \bm{A})&= f_t(\bm{W}_0,\bm{A}_0)+\nabla_{W} f_t(\bm{W}_0, \bm{A}_0)\cdot [\bm{W}-\bm{W}_0]\\
& +\nabla_{A} f_t(\bm{W}_0, \bm{A}_0)\cdot [\bm{A}-\bm{A}_0] ,\notag\\
f_t^{lin, \tau}(\bm{W}, \bm{A})&= f^\tau_t(\bm{W}_0,\bm{A}_0)+\nabla_{W} f^\tau_t(\bm{W}_0, \bm{A}_0)\cdot [\bm{W}-\bm{W}_0] \\
&+\nabla_{A} f^\tau_t(\bm{W}_0, \bm{A}_0)\cdot [\bm{A}-\bm{A}_0]. \notag
\end{aligned}
\end{equation}
We have:
\begin{lem}\label{cut}
Under the condition in Theorem \ref {mt}, with probability at least $1-\delta$,
\begin{equation}
||f^{lin}_t(\bm{W},\bm{A})-f^{lin, \tau}_t(\bm{W},\bm{A})||\leq \Theta(\frac{\sqrt{m}\tau \rho_0^{\tau}}{(1-\rho_0)^3}).
\end{equation}
\end{lem}
Then since we  set
\begin{equation}\label{tm}
\begin{aligned}
T_{max}> &\Theta(\frac{1}{\log(\frac{1}{\rho_0})} )\cdot \{ 3\log(\frac{1}{1-\rho_0})+\log b \notag \\
&+\log(\frac{1}{\epsilon} )+\frac{1}{2}\log(m)+\log T_{max}\} ,.\notag
\end{aligned}
\end{equation}
We have
\begin{equation}\label{app}
||f^{lin}_t(\bm{W},\bm{A})-f^{lin, T_{max}}_t(\bm{W},\bm{A})||\leq \epsilon/b.
\end{equation}
 Therefore $f^{lin, T_{max}}_t(\bm{W},\bm{A})$ is a good approximation for $f_t(\bm{W},\bm{A})$.

Note that from the above arguments and Lemma \ref{lr2}, we have
\begin{equation}\label{bd}
||f_t(\bm{W},\bm{A})||\leq 2b, ||\nabla_f L||\leq 2l_0\cdot(1+2b), \notag
\end{equation}
and
\begin{equation}
L(y_t, f_t(\bm{W},\bm{A}))\leq 4l_0\cdot(b+ 2b^2).
\end{equation}

\section{Proof of the Theorem \ref{mt} }\label{sk}
Theorem \ref{mt} is a direct corollary of  Theorem \ref{mt1} and \ref{mkt} below.
\begin{theo}\label{mt1} Under the condition in Theorem \ref {mt},
for any $\bm{A}^*, \bm{W}^*$ with $||\bm{A}^*-\bm{A}_0||_F, ||\bm{W}^*-\bm{W}_0||_F\leq R/\sqrt{m}\leq b\cdot T^2_{max}/\sqrt{m}$, let
\begin{equation}
\begin{aligned}
&L^k(\bm{W}_{k}, \bm{A}_{k})=\frac{1}{T}\sum_{t=1}^TL(y^k_{t},f_{t}(\bm{W}_{k}, \bm{A}_{k})),\\
&L^k_t(\bm{W}_{k}, \bm{A}_{k})=L(y^k_{t},f_{t}(\bm{W}_{k}, \bm{A}_{k})). \notag
\end{aligned}
\end{equation}
with probability at least $1-\delta$, the outputs of algorithm \ref{a1} satisfy:
\begin{equation}
\frac{1}{K}\sum_{k=0}^{K-1}L^k(\bm{W}_k, \bm{A}_k)-L^k(\bm{W}^*, \bm{A}^*)\leq \cO(\epsilon).
\end{equation}
\end{theo}

When the loss function is convex, one can easily see Theorem  \ref{mt1} follows. In our case,  the proof of Theorem \ref{mt1} is from the linearization Lemma \ref{l9}.   Lemma \ref{l9} says when  $||\bm{A}^*-\bm{A}_0||_F, ||\bm{W}^*-\bm{W}_0||_F$ are small enough,
\begin{equation}
\begin{aligned}
f_t(\bm{W},\bm{A})=\sum _{t_0=0}^{t-1} \rho^{t_0}\bm{B} (\prod_{\tau'=1}^{t_0} \bm{W})  \bm{A}X_{t-t_0}.
\end{aligned}
\end{equation}
will nearly be a linear function for $\bm{W}$ and $\bm{A}$.  This is the main  process to prove  Theorem  \ref{mt1}.

{\bf Proof of the Theorem \ref{mt1}:}

From Lemma \ref{lc}, for any $t$, $$||\nabla_W f_t||_F, ||\nabla_A f_t||_F\leq  32\frac{\sqrt{m}}{(1-\rho_0)^3}$$ when $\bm{W} \in B(\bm{W}_0, \omega)$, $\bm{A} \in B(\bm{A}_0, \omega)$. Meanwhile,
\begin{equation}
\begin{aligned}
&\bm{W}_{k+1}-\bm{W}_k=\eta \nabla_W L^k(\bm{W}_k, \bm{A}_k),\\
&\bm{A}_{k+1}-\bm{A}_k=\eta \nabla_A L^k(\bm{W}_k, \bm{A}_k).
\end{aligned}
\end{equation}

Thus for any $k\leq K$,
\begin{equation}
\begin{aligned}
&||\bm{W}_k-\bm{W}_0||_F\leq \eta K\cdot ||\nabla_W f||_F\cdot ||\nabla_f L||,\\
& ||\bm{A}_k-\bm{A}_0||_F\leq \eta K\cdot ||\nabla_A f||_F\cdot ||\nabla_f L||.
\end{aligned}
\end{equation}
In our case, from Eq (\ref{lip}) and (\ref{bd}), we have $$||\nabla_f L||\leq l_0(1+2b),$$
$$ ||\nabla_W L||_F, ||\nabla_A L||_F\leq 32\frac{\sqrt{m}}{(1-\rho_0)^3}\cdot l_0(1+2b) .$$
Thus
$||\bm{W}_k-\bm{W}_0||_F, ||\bm{A}_k-\bm{A}_0||_F\leq K\eta\cdot 32\frac{\sqrt{m}}{(1-\rho_0)^3}\cdot l_0(1+2b) \triangleq\omega.$
%For an arbitrary small $\epsilon>0$, we select $\eta= \cO( \frac{\epsilon(1-\rho_0)^4}{m})$, $K=\cO(\frac{R^2}{\epsilon m\eta})$.

From the convexity of $L$ and Lemma \ref{l9}, we have
\begin{equation}
\begin{aligned}
&L^k(\bm{W}_k, \bm{A}_k)-L^k(\bm{W}^*, \bm{A}^*)\\
&\leq \frac{1}{T}\sum_{t=1}^T\nabla_{f_t} L^k_t(\bm{W}_k, \bm{A}_k) \cdot [f_t(\bm{W}_k, \bm{A}_k)-f_t(\bm{W}^*, \bm{A}^*)],\\
&\leq  \frac{1}{T}\sum_{t=1}^T\nabla_{f_t} L^k_t(\bm{W}_k, \bm{A}_k) \cdot [\nabla_W f_t(\bm{W}_k, \bm{A}_k)\cdot [\bm{W}_k-\bm{W}^*]\\
&+\nabla_A f_t(\bm{W}_k, \bm{A}_k)\cdot [\bm{A}_k-\bm{A}^*]] + \Theta(\frac{\sqrt{m}\omega^2}{(1-\rho_0)^5})\cdot l_0(1+2b),\\
&\leq  \nabla_W L^k(\bm{W}_k, \bm{A}_k) \cdot [\bm{W}_k-\bm{W}^*]\\
&+\nabla_A L^k(\bm{W}_k, \bm{A}_k)\cdot [\bm{A}_k-\bm{A}^*]+\Theta(\frac{\sqrt{m}\omega^2}{(1-\rho_0)^5})\cdot l_0(1+2b). \notag
\end{aligned}
\end{equation}

Since $$\bm{W}_{k+1}-\bm{W}_k=-\eta \nabla_W L^k(\bm{W}_k \bm{A}_k),$$  $$\bm{A}_{k+1}-\bm{A}_k=-\eta \nabla_A L^k(\bm{W}_k, \bm{A}_k),$$ we have
\begin{align}
\frac{1}{K}&\sum_{k=0}^{K-1}\{L^k(\bm{W}_k, \bm{A}_k)-L^k(\bm{W}^*, \bm{A}^*)\} \notag\\
\leq& \frac{1}{K\eta}\sum_{k=0}^{K-1}\{\langle \bm{W}_{k}-\bm{W}_{k+1}, \bm{W}_k-\bm{W}^*\rangle \notag\\
&+ \langle \bm{A}_{k}-\bm{A}_{k+1}, \bm{A}_k-\bm{A}^*\rangle \} +\Theta(\frac{\sqrt{m}\omega^2}{(1-\rho_0)^5})\cdot l_0(1+2b),\notag\\
\leq& \frac{1}{2K\eta}\sum_{k=0}^{K-1} \{||\bm{W}_k-\bm{W}^*||^2_F -||\bm{W}_{k+1}-\bm{W}^*||^2_F \notag \\
& + ||\bm{W}_{k+1}-\bm{W}_k||^2_F + ||\bm{A}_{k}-\bm{A}^*||_F^2-||\bm{A}_{k+1}-\bm{A}^*||_F^2 \notag \\
&+ ||\bm{A}_{k+1}-\bm{A}_k||_F^2\} +\Theta(\frac{\sqrt{m}\omega^2}{(1-\rho_0)^5})\cdot l_0(1+2b), \notag \\
\leq& \frac{1}{2K\eta} \{||\bm{W}_0-\bm{W}^*||^2_F +||\bm{A}_{0}-\bm{A}^*||_F^2\}\notag \\
&+ \eta[32\frac{\sqrt{m}}{(1-\rho_0)^3}\cdot l_0(1+2b)]^2 +\Theta(\frac{\sqrt{m}\omega^2}{(1-\rho_0)^5})\cdot l_0(1+2b),\notag \\
\leq& \frac{R^2}{Km\eta}+\eta[32\frac{\sqrt{m}}{(1-\rho_0)^3}\cdot l_0(1+2b)]^2\\
&+\Theta(\frac{\sqrt{m}\omega^2}{(1-\rho_0)^5})\cdot l_0(1+2b),\notag\\
 \leq & \cO(\epsilon) +\Theta(\frac{\eta m l^2_0(1+2b)^2}{(1+\rho_0)^6})\\
 &+  \Theta([K\eta\cdot \frac{\sqrt{m}}{(1-\rho_0)^3}\cdot l_0(1+2b)]^2\cdot \frac{\sqrt{m}}{(1-\rho_0)^5})\cdot l_0(1+2b)  , \notag\\
 \leq &\cO(\epsilon)+\cO(\epsilon) \notag \\
 &+\Theta([K^2\eta^2\cdot \frac{m\sqrt{m}}{(1-\rho_0)^{11}})\cdot l_0^2(1+2b)^2\cdot l_0(1+2b)]. \notag\\
  &\leq \cO(\epsilon). \notag
 \end{align}

Meanwhile
$\frac{R}{\sqrt{m}}\leq \omega\leq \omega_0\leq \frac{1}{\rho_0}-1.$
Thus
\begin{equation}
\frac{1}{K}\sum_{k=1}^KL^k(\bm{W}_k, \bm{A}_k)-L^k(\bm{W}^*, \bm{A}^*)\leq \cO(\epsilon).
\end{equation}

Then Theorem \ref{mt1} follows. \QEDA

Based on  Theorem \ref{mt1}, to prove the main result, we need to show there exits  a $(\bm{W}^*, \bm{A}^*)$  with $||\bm{A}^*-\bm{A}_0||_F, ||\bm{W}^*-\bm{W}_0||_F\leq \cO( b\cdot T^2_{max}/\sqrt{m})$ and $||f_{t}(\bm{W}^*,\bm{A}^*))- \widetilde{y}_t|| $ small so the target lies in the  linearization  domain. In fact we have:
\begin{theo}\label{mkt}
Under the condition in Theorem \ref{mt},  with probability at least $1-\delta$, there exist $\bm{W}^*, \bm{A}^*$ with
\begin{equation}
||\bm{W}^*-\bm{W}_0||_F, ||\bm{A}^*-\bm{A}_0||_F\leq \cO(b\cdot T^2_{max}/\sqrt{m}).
\end{equation}
and
$$\frac{1}{K}\sum_{k=0}^{K-1} \frac{1}{T}\sum_{t=1}^T\{L(y^k_{t},f_{t}(\widetilde{\bm{W}}^*, \bm{A}^*, x^k))- L(y^k_{t}, \widetilde{y}^k_{t})\}\leq \cO(\epsilon).$$
\end{theo}

To prove Theorem \ref{mkt}, assume the target stable system is
\begin{equation}
\begin{aligned}
&p^i_{t}=\bm{C}p^i_{t-1}+\bm{D}x^i_t,\\
&\widetilde{y^i_t}=\bm{G}p^i_t,
\end{aligned}
\end{equation}
We construct
\begin{equation}
\begin{aligned}
&\bm{W}^*-\bm{W}_0 \\
=&\sum_{t_m=1}^{T_{max}} (\rho^{-1})^{t_m-1}  \sum_{t_1+t_2=t_m-1}\{(\prod_{\tau=1}^{t_1-1}\bm{W}_0)\}^T\bm{B}^T \bm{P}^1_{t_1}\{\bm{G}\bm{C}^{t_m-1}\bm{D} \\
&-(\rho^2)^{t_m-1}/(t_m-1)\cdot \bm{B}\{\prod_{\tau=1}^{t_m} \bm{W}_0\}\bm{A}_0 \} \bm{P}^2_{t_2-1}\bm{A}^T_0 (\prod_{\tau=1}^{t_2} \bm{W}_0), \notag
\end{aligned}
\end{equation}
\begin{equation}
\bm{A}^*-\bm{A}_0=\{(\prod_{\tau=1}^{t_m-1}\bm{W}_0)\}^T\bm{B}^T \bm{P}^1_{t_m-1}[\bm{G}\bm{D}-\bm{B}\bm{A}_0], \notag
\end{equation}

where
\begin{equation}
\begin{aligned}
&\bm{P}^1_{t_1}= \{\bm{B}(\prod_{\tau=1}^{t_1-1}\bm{W}_0)(\prod_{\tau=1}^{t_1-1}\bm{W}_0)^T\bm{B}^T\}^{-1},\\
&\bm{P}^2_{t_2}= \{\bm{A}_0^T(\prod_{\tau=1}^{t_2-1}\bm{W}_0)^T(\prod_{\tau=1}^{t_2-1}\bm{W}_0)\bm{A}_0\}^{-1}.
\end{aligned}
\end{equation}

We can  show our $\bm{W}^*$, $\bm{A}^*$  satisfying
$$||f_{t}(\bm{W}^*,\bm{A}^*))-\sum _{t_0=0}^{t-1} \bm{G} (\prod_{\tau=1}^{t_0} \bm{C})  \bm{D}X_{t-t_0}||_F\approx 0,$$
using  Lemma \ref{lr3}.
To show
$||\bm{W}^*-\bm{W}_0||_F, ||\bm{A}^*-\bm{A}_0||_F\leq \cO(b\cdot T^2_{max}/\sqrt{m})$,
we need Lemma \ref{lr2}.  The detailed proof is in Supplementary Materials.

Combing the above two theorems, we have
\begin{equation}
\frac{1}{K}\sum_{k=0}^{K-1}\frac{1}{T}\sum_{t=1}^T\{L(y^k_{t},f_{t}(\bm{W}_{k}, \bm{A}_{k}, x^k))- L(y^k_{t}, \widetilde{y}^k_{t})\}\leq \cO(\epsilon). \notag
\end{equation}
Therefore Theorem \ref{mt} follows.
\section{ Proof of Theorem \ref{gen}}
In order to prove the bound for the population risk, we need the following results for  Rademacher Complexity. These results can be found  in Proposition A.12 of \cite{allenzhu2020learning} and section 3.8 in \cite{Liang}.
\begin{theo}\label{rm}
Let $\mathcal{F}_1,\mathcal{F}_2, ... \mathcal{F}_{d_y} $ be $d_y$ classes of functions $\mathbb{R}^d\to \mathbb{R}$ and $|L(\cdot)|\leq 4l_0\cdot(b+ 2b^2)$ be a bounded, $l_0(1+2b)$  Lipschitz function. For $N$ samples $x_i$ i.i.d. drawn from a distribution $\dataset$, with probability at least $1-\delta$, we have
\begin{equation}
\begin{aligned}
\sup_{f_1\in\mathcal{F}_1,... f_{d_y}\in\mathcal{F}_{d_y}} &|\mathbb{E}_{x\sim \dataset} [L(f_1(x), ... f_{d_y}(x))]\\
&-\frac{1}{N}\sum_{i=1}^NL(f_1(x_i), ... f_{d_y}(x_i))|\\
&\leq  \Theta(l_0(1+2b)\cdot \sum_{r=1}^{d_y} \mathscr{R}(\mathcal{F}_r))\\
&+4\frac{4l_0\cdot(b+ 2b^2)\sqrt{\log(1/\delta)}}{\sqrt{N}}. 
\end{aligned}
\end{equation}
where  $\mathscr{R}(\mathcal{F}))$ is defined as :
\begin{equation}
 \mathscr{R}(\mathcal{F}))=\mathbb{E}_{\xi\sim \{\pm 1\}^N} [\sup_{f\in \mathcal{F}}\frac{1}{N}\sum_i^N \xi_i f(x_i)].
\end{equation}
\end{theo}
Meanwhile, for linear functions,  Rademacher Complexity  is easy to be calculated:
\begin{theo}\label{lg}(Proposition A.12 in  \cite{allenzhu2020learning})

Let $\mathcal{F}$ be the function class $\{x\to  f_0(x)+ \langle w, x\rangle\in \mathbb{R}\}$ with $f_0$ a  fixed function and $||w||\leq B$, then
\begin{equation}
 \mathscr{R}(\mathcal{F}))\leq \frac{B}{\sqrt{N}}.
\end{equation}
\end{theo}

Now we will estimate the gap between the  empirical and population loss:
\begin{equation}
\begin{aligned}
&|\mathbb{E}_{x,y\sim\dataset} \frac{1}{T}\sum_{t=1}^T\{L(y_{t},f_{t}(\bm{W}_{k}, \bm{A}_{k}, x))- L(y_{t}, \widetilde{y}_{t})\}\\
& -\frac{1}{K}\sum_{k=0}^{K-1}\frac{1}{T}\sum_{t=1}^T\{L(y^k_{t},f_{t}(\bm{W}_{k}, \bm{A}_{k}),x^k)- L(y^k_{t}, \widetilde{y}^k_{t})\}|.\notag
\end{aligned}
\end{equation}
Firstly, from Eq (\ref{lip}) and  (\ref{app}),
\begin{equation}
\begin{aligned}
|L(y_{t},f_{t}(\bm{W}_{k}, \bm{A}_{k}))&-L(y_{t},f^{lin, T_{max}}_{t}(\bm{W}_{k}, \bm{A}_{k}))|\leq \cO(\epsilon), \notag
\end{aligned}
\end{equation}
and  for all $k\leq K$, $$||\bm{W}_{k}-\bm{W}_{0}||_F, ||\bm{A}_{k}-\bm{A}_{0}||_F\leq \omega,$$
where $\omega =\cO(\frac{K\eta\sqrt{m} }{(1-\rho_0)^3}\cdot l_0 (1+2b)).$ Let $e_j$ be the $j$-th orthogonal basis of $\mathbb{R}^{d_y}$.  We only need to consider the Rademacher complexity of the function class:
 \begin{equation}
 \begin{aligned}
 \{& x\to  e_j^Tf^{lin, T_{max}}_{t}(\bm{W}_0+\Delta \bm{W} , \bm{A}_{0}+\Delta\bm{A}, x)|\\
 &\ ||\Delta \bm{W}||_F, ||\Delta \bm{A}||_F\leq \omega \}.
 \end{aligned}
 \end{equation}
And this is a class of linear functions.  We can write it as $x\to \langle F_0, x\rangle+\langle F, x\rangle$ with $F_0$ fixed.  Apply Lemma \ref{lc}.  We have
\begin{equation}
\begin{aligned}
&||F||_2\leq 32\frac{\sqrt{m}}{(1-\rho_0)^3}\omega\\
&\leq [32\frac{\sqrt{m}}{(1-\rho_0)^3}] \cdot [K\eta\cdot 32\frac{\sqrt{m}}{(1-\rho_0)^3}\cdot l_0(1+2b)].
\end{aligned}
\end{equation}

Then  combing all the above results,  Theorem \ref{rm} and Theorem  \ref{lg},  we have
\begin{equation}
\begin{aligned}
&|\mathbb{E}_{x,y\sim\dataset} \frac{1}{T}\sum_{t=1}^T\{L(y_{t},f_{t}(\bm{W}_{k}, \bm{A}_{k}, x))- L(y_{t}, \widetilde{y}_{t})\}\\
& -\frac{1}{K}\sum_{k=0}^{K-1}\frac{1}{T}\sum_{t=1}^T\{L(y^k_{t},f_{t}(\bm{W}_{k}, \bm{A}_{k}),x^k)- L(y^k_{t}, \widetilde{y}^k_{t})\}|\\
&\leq  \Theta( l_0(1+2b) [\frac{\sqrt{m}}{(1-\rho_0)^3}]) \\
&\cdot \Theta([K\eta\cdot\frac{\sqrt{m}}{(1-\rho_0)^3}\cdot l_0(1+2b)]/ \sqrt{K})\\
&+\cO(\frac{l_0\cdot(b+ 2b^2)\sqrt{\log(1/\delta)}}{\sqrt{K}})\\
&\leq  \cO(l_0^2(1+2b)^2\frac{\eta m \sqrt{K}}{(1-\rho_0)^6})\\
&+\cO(\frac{l_0\cdot(b+ 2b^2)\sqrt{\log(1/\delta)}}{\sqrt{K}})\\
&\leq \cO(\epsilon).
\end{aligned}
\end{equation}
with probability at least $1-\delta$. Our claim follows.
\section*{Acknowledgement}
This research was funded by the Foundation item: National Natural Science Foundation of China (U1936215). Project name: Methodologies and Key Technologies of Intelligent Detection and Tracing of APT Attacks.

\section{Conclusion}
We provided the first theoretical guarantee on learning linear RNNs with Gradient Descent. The required width in hidden layers does not rely on the length of the input sequence and only depends on the transition matrix parameter $\rho_C$. Under this condition, we showed that SGD can provably learn  any stable linear system with  transition matrix $\bm{C}$ satisfying $||\bm{C}^k||\leq \cO(\rho_C^k), k\in \mathbb{N}$, using  $poly(\frac{1}{1-\rho_C}, \epsilon^{-1})$ many iterations and $poly(\frac{1}{1-\rho_C},  \epsilon^{-1})$ many samples. In this work we found a suitable random initialization which is available to optimize using gradient descent. This
 solves an open problem in System Identification and answers why SGD is available to optimize  RNNs in practice. We hope this result can provide some insights for learning stable nonlinear dynamic systems using recurrent neural networks in deep learning.

\bibliography{ijcai22}
\bibliographystyle{IEEEtran}

{\appendix
%\onecolumn
%{\bf \LARGE Supplementary Materials}
\section*{Lemma}
The following concentration inequality is a standard result for subexponential distribution, which will be use many times.
\begin{lem}Standard Concentration inequality for Chi-square distribution:

Let $i=1,2,3,...m, v\in \mathbb{N}$ and $A_i\sim \chi^2(v)$ be $m$ i.i.d Chi-square distribution. Then with probability at least $1-exp(-\Omega(m\epsilon))$:
$$|\sum_{i=1}^m\frac{1}{m} A_i-\mathbb{E}\chi^2(v)|\leq \epsilon.$$
\end{lem}
As a corollary, let $\bm{A}\in \mathbb{R}^{m\times n}$ is a Gaussian random matrix, and the element $A_{i,j}\sim \mathcal{N}(0,\frac{1}{m})$.   We can see for a fixed $v$ with $||v||=1$, with probability at least $1-exp(-\Omega(m\epsilon^2))$, 
$$|\ ||\bm{A}v||-1|\leq \epsilon.$$
\section*{Proof of Lemma \ref{lc}}
Firstly,
\begin{equation}
\begin{aligned}
||\sum_{t=\tau}^{\infty} \rho^{t}\cdot \bm{B} (\prod_{\tau=1}^{t} \bm{W}) \bm{Q} Z_t|| \leq \sum_{t=\tau}^{\infty} \rho^{t}||\bm{B} (\prod_{\tau=1}^{t} \bm{W}) \bm{Q} Z_t||.\\
\end{aligned}
\end{equation}
Meanwhile, thanks to (\ref{wr}), for all $k$ in $\mathbb{N}$,  $||\rho^k\bm{W}^k||\leq 2\sqrt{k}\rho_0^k$, 
$$\rho^{k}||\bm{B} (\prod_{\tau=1}^{k} \bm{W})||\leq 4\sqrt{m}\sqrt{k}\cdot \rho_0^k ||\bm{Q}||\cdot ||Z_t||.$$

 Thus \begin{equation}
\begin{aligned}
&||\sum_{t=\tau}^{\infty} \rho^{t}\cdot \bm{B} (\prod_{\tau=1}^{t} \bm{W}) \bm{Q} Z_t||\\
&\leq4\frac{\tau\sqrt{m}(\rho_0)^{\tau}}{(1-\rho_0)^2}||\bm{Q}||
 \end{aligned}
\end{equation} The first part of Lemma \ref{lc} follows.

Now we consider
\begin{equation}
||\sum_{t_0=\tau}^\infty \sum_{t_1+t_2=t_0}  \rho^{t_0}\bm{B}(\prod_{\tau=1}^{t_1-1}\bm{W})  \bm{Q}  (\prod_{\tau=1}^{t_2-1} \bm{W})  \bm{A}Z_{t_0}||.
\end{equation}

Note that
\begin{equation}
\begin{aligned}
&||\rho^{t_0}\bm{B}(\prod_{\tau=1}^{t_1-1}\bm{W})  \bm{Q}  (\prod_{\tau=1}^{t_2-1} \bm{W})  \bm{A}Z_{t_0}||\\
&\leq  8\sqrt{m} \rho_0^{t_0}t_0||\bm{Q}||.
\end{aligned}
\end{equation}
Thus
\begin{equation}
\begin{aligned}
||\sum_{t_0=\tau}^\infty \sum_{t_1+t_2=t_0} & \rho^{t_0}\bm{B}(\prod_{\tau=1}^{t_1-1}\bm{W})  \bm{Q}  (\prod_{\tau=1}^{t_2-1} \bm{W})  \bm{A}Z_{t_0}||\\
&\leq 32\frac{\sqrt{m}(\rho_0)^{\tau}\tau^2}{(1-\rho_0)^3}||\bm{Q}||.
\end{aligned}
\end{equation}
Our results follow.

\QEDA

\section*{ Proof of Lemma \ref{lr}}

For a fixed $v_2$, with probability at least $1-exp(-\Omega(m\epsilon^2))$, $1-\epsilon\leq||\bm{A}_0v_2||\leq 1+\epsilon$.

Taking the $\epsilon-net$ in $\mathbb{R}^{d}$, there is a set $\mathcal{N}$ with $|\mathcal{N}|\leq (3/\epsilon)^d$. For any vector $v$ in $\mathbb{R}^{d}$ with $||v||=1$, there is a vector $v'$ satisfying $||v-v'||\leq \epsilon$. Thus with probability at least $1-(3/\epsilon)^dexp(-\Omega(m\epsilon^2))=1-exp(-\Omega(m\epsilon^2))$, for any vector $v_2$ in $\mathbb{R}^{d}$ with $||v_2||=1,$
\begin{equation}
1-\epsilon- (1+\epsilon)\epsilon\leq ||\bm{A}_0v_2||\leq (1+\epsilon)^2.
\end{equation}
Thus
\begin{equation}
1-2\epsilon-\epsilon^2 \leq ||\bm{A}_0v_2||\leq 1+2\epsilon +\epsilon^2.
\end{equation}

This  is the same for $||\bm{W}_0\bm{A}_0v_2||$. With probability at least $1-exp(-\Omega(m\epsilon^2))$
\begin{equation}\label{singlar}
1-\epsilon\leq||\bm{W}_0\bm{A}_0v_2||\leq 1+\epsilon.
\end{equation}
Let $h_t=\prod_{t_0=1}^t \bm{W}_0\bm{A}_0v_2$. When $t>1$, we consider the Gram-Schmidt orthogonalization.

Let  $\bm{U}_t$ be the Gram-Schmidt orthogonalization matrix as:
\begin{equation}
\bm{U}_t=GS(h_0, h_1,...h_t).
\end{equation}
We have
\begin{equation}\label{ep}
\begin{aligned}
h_t=&\bm{W}_0h_{t-1}= \bm{W}_0\bm{U}_{t-2}\bm{U}_{t-2}^Th_{t-1}\\
& +\bm{W}_0[\bm{I}-\bm{U}_{t-2}\bm{U}_{t-2}^T]h_{t-1},\\
=&\begin{bmatrix}
\bm{W}_0\bm{U}_{t-2}, & \frac{\bm{W}_0[\bm{I}-\bm{U}_{t-2}\bm{U}_{t-2}^T]h_{t-1}}{||[\bm{I}-\bm{U}_{t-2}\bm{U}_{t-2}^T]h_{t-1}||}
\end{bmatrix}
\\&\cdot
\begin{bmatrix}
\bm{U}_{t-2}^Th_{t-1}\\
||[\bm{I}-\bm{U}_{t-2}\bm{U}_{t-2}^T]h_{t-1}||
\end{bmatrix},\\
=&\begin{bmatrix}
M_1, & M_2
\end{bmatrix}
\cdot
\begin{bmatrix}
z_1\\
z_2
\end{bmatrix}.
\end{aligned}
\end{equation}
Using this expression, the entries of  $M_1=\bm{W}_0\bm{U}_{t-2}$ and $M_2=\frac{\bm{W}_0[\bm{I}-\bm{U}_{t-2}\bm{U}_{t-2}^T]h_{t-1}}{||[\bm{I}-\bm{U}_{t-2}\bm{U}_{t-2}^T]h_{t-1}||}$ are i.i.d from $\mathcal{N}(0,\frac{1}{m})$. Therefore for fixed $z_1, z_2$, with probability at least $1-exp(-\Omega(m\epsilon^2))$, $$(1-\epsilon) \sqrt{z_1^2+z_2^2}\leq ||h_t||\leq (1+\epsilon) \sqrt{z_1^2+z_2^2}.$$  Take the $\epsilon-net$ for $z_1. z_2$. The sizes of $\epsilon-net$ for $z_1. z_2$ are $(3/\epsilon)^{d\cdot (t+1)}$ and  $(3/\epsilon)^{1}$. Thus with probability at least $1-(3/\epsilon)^dexp(-\Omega(m\epsilon^2))=1-exp(-\Omega(m\epsilon^2)),$   $$(1-\epsilon) \sqrt{z_1^2+z_2^2}\leq ||h_t||\leq (1+\epsilon) \sqrt{z_1^2+z_2^2}$$ for any $z_1, z_2$.

Therefore we have
\begin{equation}\label{tro}
(1-\epsilon)^t \leq||\prod_{t_0=1}^{t}\bm{W}_0\bm{A}_0v_2||\leq (1+\epsilon)^t.
\end{equation}
for any $0\leq t\leq \tau$.
Set $\epsilon=\frac{0.01}{\tau}$. The theorem for $||\bm{W}^t_0\bm{A}_0v_2||$ follows. For $||\frac{\sqrt{d_y}}{\sqrt{m}}(\bm{W}^T_0)^t\bm{B}^Tv_1||$, the proof is the same.

This proves Lemma \ref{lr}.

\section*{ Proof of Lemma \ref{sp} }

Now we will prove  Lemma \ref{sp}. Similar to (\ref{tro}),  for fixed $v\in \mathbb{R}^m$,  let $L=\Theta(\frac{\sqrt{m}}{\log m})\in \mathbb{N}$ . With probability at least $1-Lexp(-\Omega (m/L^2))$,  for all  $0<t\leq L$, we have
\begin{equation}
(1-\frac{1}{L})^t||v|| \leq||\prod_{t_0=1}^{t}\bm{W}_0v||\leq (1+\frac{1}{L})^t||v||.
\end{equation}
 Then  for fixed normalized  orthogonal basis  $v_i$, $i=1,2,..m$, with probability at least $1-Lmexp(-\Omega ( m/L^2))$, for all  $0<t\leq L$, $i=1, 2, ... m$,
\begin{equation}
||\bm{W}^t_0v_i||\leq (1+1/L)^t||v_i||.
\end{equation}
Any $v\in \mathbb{R}^m$ can be write as $v=\sum_{i=1}^m a_i v_i$. When $||v||=1$, we have $\sum_i |a_i|\leq \sqrt{m}$.

Therefore with probability at least $1-Lmexp(-\Omega ( m/L^2))=1-exp(-\Omega ( m/L^2))$, for all  $0<t\leq L$, any $v$
\begin{equation}
||\bm{W}^t_0v||\leq \sqrt{m}(1+1/L)^t||v||.
\end{equation}
Since $L=\Theta(\frac{\sqrt{m}}{\log m})$, $L^2>\sqrt{m}$, we have
\begin{equation}
||\bm{W}^t_0||\leq L^2(1+1/L)^t.
\end{equation}

Note that  $1/\rho_1= 1+10\cdot \frac{\log^2 m}{\sqrt{m}} >1+\log L\cdot \frac{\log m}{\sqrt{m}}$  so we can set $L=\Theta(\frac{\sqrt{m}}{\log m})\in \mathbb{N}$ lager than an absolute constant such that
\begin{equation}
\begin{aligned}
||\bm{W}^L_0||&\leq L^2(1+1/L)^L\leq e^{2\log L}\cdot e\\
&=e^{1+2\log L}\leq (1+10\cdot \frac{\log^2 m}{\sqrt{m}})^{(\frac{\sqrt{m}}{\log^2 m})\cdot 100\cdot \log L}\\
&\leq (1/\rho_1)^{(\frac{\sqrt{m}}{\log m})\cdot 100\cdot \frac{\log L}{\log m}}\\
& \leq  (1/\rho_1)^{L/2}.
\end{aligned}
\end{equation}
And for $k\leq L$
\begin{equation}
\begin{aligned}
||\bm{W}^k_0||&\leq L^2(1+1/L)^L \leq  (1/\rho_1)^{L/2}.
\end{aligned}
\end{equation}
If $s>L$, we can write it as $s=k\cdot L +r$ and $k, r\in \mathbb{N}$, $r\leq L$. Then
\begin{equation}
||\bm{W}^s_0||\leq ||\bm{W}^L_0||^k \cdot ||\bm{W}^r_0||\leq \rho_1^{-Lk/2}\cdot \rho_1^{-L/2}\leq \rho_1^{-s} .
\end{equation}
Thus we have
\begin{equation}
||\bm{W}^s_0||\leq \rho_1^{-s}.
\end{equation}

For $k \leq 2L$, note that for any unit vector $v\in \mathbb{R}^m$, we can write $v=\sum_{i=1}^ku_i$, where $u_i$ has at most $2m/k$ non-zero coordinates and the intersection of non-zero coordinates sets for $u_i$ and $u_j$ are empty when $i\neq j$. In space $R^{2m/k}$, we can use the $\epsilon$-net argument which says with probability at least $1-exp(-\Omega(m/k))$, for any $v\in \mathbb{R}^{2m/k}$,
$$||W_0^kv||\leq 2||v||.$$
Thus with probability at least $1-4L^2exp(-\Omega(m/L))$, for all $k\leq 2L$, $||W_0^k||\leq 2\sqrt{k}$.
\section*{ Proof of Lemma \ref{lr2}}
Note that  for a fixed $v\in \mathbb{R}^m$,
\begin{equation}
||\bm{B}v||
\end{equation}
 is a Gaussian random variable with variance $||v||^2$. Meanwhile, let $v_i$, $i=1,...d$  be the orthogonal basis in $\mathbb{R}^d$. We only need to bound  $\tau\cdot d$ many vectors $B\bm{W}^t_0\bm{A}_0v_i$. 

Thus for any $t\leq \tau$, and any $v_i$, with probability at least $1-(\tau\cdot d)\cdot exp(-\Omega(m\epsilon^2))$, we have 
 \begin{equation}
||\prod_{t_0=1}^{t}\bm{W}_0\bm{A}_0v_i||\leq (1+\epsilon)^t.
\end{equation}
Thus if $m>\Omega(\log(\tau \cdot d/\delta)$, with probability at least $1 - \delta$,
 \begin{equation}
 ||\bm{B}\bm{W}^t_0\bm{A}_0||\leq  \sqrt{d\log(\tau\cdot d/\delta)}).
 \end{equation}
 Lemma \ref{lr2} follows,
\section*{ Proof of Lemma \ref{lr3}}
Consider
\begin{equation}\label{ev}
|u^T\bm{A}\{(\prod_{\tau=1}^{t}\bm{W}_0)^T\}\{(\prod_{\tau=1}^{t'}\bm{W}_0)\}\bm{A}v|,
\end{equation}
with $||u||=||v||=1$.
Before we study the  properties of the above equation, we need some 
 very useful results in random matrices theory.
\begin{lem}(Lemma 5.36 in \cite{vershynin2011introduction})
Consider a matrix $\bm{B}$ and $\delta>0$. Let $s_{min}(\bm{B}), s_{max}(\bm{B})$ be the smallest and the largest singular values of $B$ which satisfy
\begin{equation}
    1-\delta \leq s_{min}(\bm{B})\leq s_{max}(\bm{B})\leq 1+\delta.
\end{equation}
Then 
\begin{equation}
    ||\bm{B}^T\bm{B}-\bm{I}||\leq 3\max(\delta,\delta^2).
\end{equation}
\end{lem}
combining this lemma with (\ref{singlar}), we have the following result.
\begin{lem}\label{ind}
For any $t\leq \tau$, let  $m^{1/2}>\tau$, $\bm{F}=\bm{W}^t\bm{A}$. With probability at least $1-\exp(-\Omega(\log^2{m}))$, we have
    \begin{equation}
    ||\bm{F}^T\bm{F}-\bm{I}||\leq \log m/\sqrt{m}.
\end{equation}
\end{lem}

Meanwhile, let $x\in \mathbb{R}^d$, $y=\bm{A}x\in \mathbb{R}^m$. We have:
\begin{equation}
    y^T\bm{W}_0y= \bm{w'}^T\bm{A}x,
\end{equation}
where $w'\in \mathbb{R}^m$ is a random vector and every entry of which is i.i.d drawn from $N(0,||y||^2)$.
Thus with probability at least $1-exp(-\Omega(\log^2m))$, we have
\begin{equation}\label{qu}
    |y^T\bm{W}_0y|\leq \log m/\sqrt{m}\cdot ||y||^2.
\end{equation}

Now let $i,j \in [d]$  be the two indexes.
Let $e_1, e_2, ... e_d$ be the orthonormal basis in $\mathbb{R}^d$.
We can write (\ref{ev}) into a linear combination of the below terms.
\begin{equation}
|e_i^T\bm{A}^T\{(\prod_{\tau=1}^{t}\bm{W}_0)^T\}\{(\prod_{\tau=1}^{t'}\bm{W}_0)\}\bm{A}e_j|,
\end{equation}

For any $i\neq j$, we set
\begin{equation}
z_{i,0}=\bm{A}e_i, z_{j,0}=\bm{A}e_j.
\end{equation}
And
\begin{equation}
z_{i,t}= \{\prod_{\tau=1}^{t}\bm{W}_0\}\bm{A} e_i.
\end{equation}
As shown in the last section, when $t\leq \tau$, with probability at least $1-\tau exp(-\Omega ( m/\tau^2))$.

\begin{equation}\label{er0}
||z_{i,t}||\leq (1+1/100\tau)^t.
\end{equation}

Now we  consider Gram-Schmidt orthonormal matrix
\begin{equation}
Z_{j,t}=GS(z_{i,0}, z_{j,0}, z_{i,1},z_{j,1},... z_{i,t}, z_{j,t}).
\end{equation}
It has at most $2(t+1)$ columns.

In order to prove the lemma, we consider $||Z_{j,t}^Tz_{i,t}||$ using induction.
\begin{comment}
    change
\end{comment}
When $t=0$ and $t=1$, from Lemma \ref{ind} and \ref{qu}, we know with probability at least $1-exp(-\Omega(\log^2m))$,  $||Z_{j,t}^Tz_{i,t}||\leq 2\log m /\sqrt{m} $.

For $t+1$,
\begin{equation}
\begin{aligned}
Z_{j,t+1}^Tz_{i,t+1}=&Z_{t+1}^T\bm{W}_0(I-Z_{j,t}Z_{j,t}^T) z_{i,t}\\
& +Z_{j,t+1}^T\bm{W}_0 Z_{j,t}Z_{j,t}^T z_{i,t}.
\end{aligned}
\end{equation}
Note that thanks to the  Gram-Schmidt orthogonalization, the elements in matrix $\bm{W}_0(I-Z_{j,t}Z_{j,t}^T)$ are  independent of $ Z_{j,t+1}$.
Then  with probability at least $1-exp(-\Omega(\log^2 m ))$,
\begin{equation}
\begin{aligned}
&||Z_{j,t+1}^T\bm{W}_0(I-Z_{j,t}Z_{j,t}^T) z_{i,t}||\\
&\leq ||Z_{j,t+1}^T||\cdot ||\frac{\bm{W}_0(I-Z_{j,t}Z_{j,t}^T) z_{i,t}}{||(I-Z_{j,t}Z_{j,t}^T) z_{i,t}||}||\cdot ||(I-Z_{j,t}Z_{j,t}^T) z_{i,t}||,\\
&\leq ||Z_{j,t+1}^T||\cdot 2\log m /\sqrt{m}\cdot ||(I-Z_{j,t}Z_{j,t}^T) z_{i,t}||,\\
&\leq  2\log m /\sqrt{m}\cdot ||(I-Z_{j,t}Z_{j,t}^T) z_{i,t}||\\
&\leq 2\log m /\sqrt{m}\cdot ||z_{i,t}||\\
&\leq  2\log m /\sqrt{m}\cdot \log m.
\end{aligned}
\end{equation}
As for $Z_{j,t+1}^T\bm{W}_0 Z_{j,t}Z_{j,t}^T z_{i,t}$, firstly note that the entries of matrix $\bm{W}_0 Z_{j,t}$ are i.i.d from $N(0,\frac{1}{m})$.  When $Z_{j,t}^T z_{i,t}$ is fixed,  with probability at least $1-exp(-\Omega(m/\tau^2)$,
\begin{equation}
\begin{aligned}
&||Z_{j,t+1}^T\bm{W}_0 Z_{j,t}Z_{j,t}^T z_{i,t}||\leq ||\bm{W}_0 Z_{j,t}Z_{j,t}^T z_{i,t}||\\
&\leq (1+\frac{1}{100\tau}) ||Z_{j,t}^T z_{i,t}||.
\end{aligned}
\end{equation}
Meanwhile, there are at most $exp(\cO(t+1))$ fixed vectors forming an $\epsilon$-net for $Z_{j,t}^T z_{i,t}$. Let  $m^{1/2}>\tau$.  With probability at least $1-exp(-\Omega(m/\tau^2))$,
\begin{equation}
||Z_{j,t+1}^T\bm{W}_0 Z_{j,t}Z_{t}^T z_{i,t}||\leq (1+\frac{1}{50\tau}) ||Z_{j,t}^T z_{i,t}||.
\end{equation}
Therefore
\begin{equation}
\begin{aligned}
||Z_{j,t+1}^T z_{i,t+1}||&=||Z_{j,t+1}^T\bm{W}_0 z_{i,t}||\\
&\leq (1+\frac{1}{50\tau}+2\frac{\log^2 m} {\sqrt{m}})^{t+1}\cdot (2\frac{\log m} {\sqrt{m}})
\end{aligned}
\end{equation}
When $m$ is large enough such that $\frac{\sqrt{m}}{2\log^2 m} >50\tau$, we have 
$$(1+\frac{1}{50\tau}+2\frac{\log^2 m} {\sqrt{m}})^{t+1}\leq 3,$$
Thus \begin{equation}\label{er1}
\begin{aligned}
||Z_{j,t+1}^T z_{i,t+1}||\leq 6\frac{\log m} {\sqrt{m}}.
\end{aligned}
\end{equation}
As a corollary, this says suppose $||u|| =||v||=1, u^Tv=0$, for all $k\leq \tau$,
$$ |u^T\bm{W}^kv|\leq 6\frac{\log m} {\sqrt{m}}.$$

Without loss of generality, we assume $t>t'$. 
\begin{equation}
\begin{aligned}
&|e_j^T\bm{A}^T\{(\prod_{\tau=1}^{t}\bm{W}_0)^T\}\{(\prod_{\tau=1}^{t'}\bm{W}_0)\}\bm{A}e_i|\\
&=[z_{j,t'}]^T\bm{W}_0^{t-t'}z_{i,t'}\\
&=[z_{j,t'}]^T\bm{W}_0^{t-t'}(I-Z_{j,t'}Z_{j,t'}^T)z_{i,t'}\\
&+[z_{j,t'}]^T\bm{W}_0^{t-t'}Z_{j,t'}Z_{j,t'}^Tz_{i,t'}\\
&\leq||z_{j,t'}||\cdot ||\bm{W}_0^{t-t'}||\cdot 6\frac{\log m} {\sqrt{m}}\\
&+||z_{j,t'}||\cdot 6\frac{\log m} {\sqrt{m}} \cdot ||z_{i,t'}||\\
&\leq 24\tau\frac{\log m} {\sqrt{m}}.
\end{aligned}
\end{equation}
Now we consider the case $i=j$, 
\begin{equation}
\begin{aligned}
&|e_i^T\bm{A}^T\{(\prod_{\tau=1}^{t}\bm{W}_0)^T\}\{(\prod_{\tau=1}^{t'}\bm{W}_0)\}\bm{A}e_i|\\
&=[z_{i,t}]^Tz_{i,t'}.
\end{aligned}
\end{equation}
Since 
\begin{equation}
\begin{aligned}
||Z_{j,t+1}^T z_{i,t+1}||\leq 6\frac{\log m} {\sqrt{m}},
\end{aligned}
\end{equation}
and \begin{equation}
Z_{j,t}=GS(z_{i,0}, z_{j,0}, z_{i,1},z_{j,1},... z_{i,t}, z_{j,t}),
\end{equation} we have
\begin{equation}
\begin{aligned}
&|[z_{i,t}]^Tz_{i,t'}|\leq 6\frac{\log m} {\sqrt{m}}.
\end{aligned}
\end{equation}
Thus for any $u,v$,
\begin{equation}
\begin{aligned}
&|u^T\bm{A}^T\{(\prod_{\tau=1}^{t}\bm{W}_0)^T\}\{(\prod_{\tau=1}^{t'}\bm{W}_0)\}\bm{A}v|\\
&\leq 24\tau\frac{d^2\log m} {\sqrt{m}}.
\end{aligned}
\end{equation}
There is a similar argument for $$\bm{B}^T (\prod_{\tau=1}^{t'} \bm{W}_0)(\prod_{\tau=1}^{t} \bm{W}_0)^T\bm{B}.$$
The theorem follows.

\QEDA

\section*{ Proof of Lemma \ref{l9} and \ref{cut}}

Firstly, note that
\begin{equation}
\begin{aligned}
f_t(\bm{W},\bm{A})&=\bm{B}\bm{A}X_t+ \rho\cdot \bm{B}\bm{W}\bm{A}X_{t-1}+...\\
&+  \rho^{t-1}\cdot \bm{B} (\prod_{\tau=1}^{t-1} \bm{W})  \bm{A}X_{1}.
\end{aligned}
\end{equation}
We have
\begin{equation}
\begin{aligned}
& \nabla_W f_t(\bm{W},\bm{A})\cdot \bm{Z} =\bm{B}\bm{Z}\bm{A}X_{t-1}+...\\
 &+  \rho^{t-1}\cdot\sum_{t_1+t_2=t-1, t_1,t_2>0} \bm{B}(\prod_{\tau=1}^{t_1-1}\bm{W})  \bm{Z}  (\prod_{\tau=1}^{t_2-1} \bm{W})  \bm{A}X_{1},\\
&=\sum_{t_1+t_2+t_0=t, t_0, t_1, t_2>0}  \rho^{t-t_0}\\
&\cdot\bm{B}(\prod_{\tau=1}^{t_1-1}\bm{W})  \bm{Z}  (\prod_{\tau=1}^{t_2-1} \bm{W})  \bm{A}X_{t_0}.\\
\end{aligned}
\end{equation}
Meanwhile,
\begin{equation}
\begin{aligned}
\nabla_A f_t(\bm{W},\bm{A})\cdot\bm{Z}=&\bm{B}Z X_t+ \rho\cdot \bm{B}\bm{W}\bm{Z}X_{t-1}+...\\
&+ \rho^{t-1}\cdot\bm{B} (\prod_{\tau=1}^{t-1} \bm{W})  \bm{Z}X_{1}.
\end{aligned}
\end{equation}

Let
\begin{equation}
R_{t_0}(\bm{W}, \bm{A})= \rho^{t-t_0}\cdot \bm{B} (\prod_{\tau=1}^{t-t_0} \bm{W})  \bm{A}X_{t_0}.
\end{equation}

Note that from (\ref{wr}), $||\rho^t \bm{W}^t||\leq 2\sqrt{t}\rho_0^{t}$ for all $t\in \mathbb{N}$. We have
\begin{equation}
\begin{aligned}
&||R_{t_0}(\bm{W}', \bm{A}')-R_{t_0}(\bm{W}, \bm{A})\\
&- \nabla_W R_{t_0}(\bm{W}, \bm{A})\cdot [\bm{W}'-\bm{W}],\\
& -\nabla_A R_{t_0}(\bm{W}, \bm{A})\cdot [\bm{A}'-\bm{A}]||_F,\\
&\leq \sum_{i+j=t-t_0-1}||\bm{B}||\cdot \rho^{t-t_0}\cdot ||\bm{W}^{i}||\cdot ||\bm{W}^{j}||\\
&\cdot ||\bm{W}'- \bm{W}||\cdot ||\bm{A'}-\bm{A}||\\
&+\sum_{i+j+k=t-t_0-2}||\bm{B}||\cdot \rho^{t-t_0}\cdot ||\bm{W}^{i}||\cdot||\bm{W}^{j}||\cdot||\bm{W}^{k}||\\
&\cdot ||\bm{W}'- \bm{W}||^2\cdot ||\bm{A}||\\
&\leq 2\sqrt{m} (t-t_0)\rho_0^{t-t_0}\cdot 4(t-t_0)\cdot \omega^2\\
&+ 2\sqrt{m} (t-t_0)^2\rho_0^{t-t_0}\cdot 8(t-t_0)^2\cdot \omega^2\cdot 2\\
&\leq 32\sqrt{m} (t-t_0)^4\omega^2\rho_0^{t-t_0}.
\end{aligned}
\end{equation}

Thus
\begin{equation}
\begin{aligned}
&||f_t(\bm{W}',\bm{A}')-f_t(\bm{W},\bm{A})-\langle \nabla_{W} f_t(\bm{W}, \bm{A}), \bm{W}'-\bm{W}\rangle \\
&+\langle \nabla_{A} f_t(\bm{W}, \bm{A}), \bm{A}'-\bm{A}\rangle ||_F,\\
&\leq \sum_{t}32\sqrt{m} (t-t_0)^4\omega^2\rho_0^{t-t_0}\\
&\leq 768\frac{\sqrt{m}\omega^2}{(1-\rho_0)^5}.
\end{aligned}
\end{equation}
Lemma \ref{l9} follows.
\QEDA

\section{Proof of Lemma \ref{cut}}

From the definition we have
\begin{equation}
\begin{aligned}
f_t^{lin}(\bm{W}, \bm{A})= &f_t(\bm{W}_0,\bm{A}_0)+\nabla_{W} f_t(\bm{W}_0, \bm{A}_0)\cdot [\bm{W}-\bm{W}_0]\\
& +\nabla_{A} f_t(\bm{W}_0, \bm{A}_0)\cdot [\bm{A}-\bm{A}_0],\\
f_t^{lin, \tau}(\bm{W}, \bm{A})= &f^\tau_t(\bm{W}_0,\bm{A}_0)+\nabla_{W} f^\tau_t(\bm{W}_0, \bm{A}_0)\cdot [\bm{W}-\bm{W}_0]\\
& +\nabla_{A} f^\tau_t(\bm{W}_0, \bm{A}_0)\cdot [\bm{A}-\bm{A}_0],
\end{aligned}
\end{equation}
and
\begin{equation}
\begin{aligned}
 &\nabla_W f_t(\bm{W},\bm{A})\cdot \bm{Z} =\bm{B}\bm{Z}\bm{A}X_{t-1}+...\\
 &+  \rho^{t-1}\cdot\sum_{t_1+t_2=t-1, t_1,t_2>0} \bm{B}(\prod_{\tau=1}^{t_1-1}\bm{W})  \bm{Z}  (\prod_{\tau=1}^{t_2-1} \bm{W})  \bm{A}X_{1}\\
&=\sum_{t_1+t_2+t_0=t, t_0, t_1, t_2>0}  \rho^{t-t_0}\cdot\bm{B}(\prod_{\tau=1}^{t_1-1}\bm{W})  \bm{Z}  (\prod_{\tau=1}^{t_2-1} \bm{W})  \bm{A}X_{t_0}.\\
\end{aligned}
\end{equation}
Meanwhile
\begin{equation}
\begin{aligned}
\nabla_A f_t(\bm{W},\bm{A})\cdot\bm{Z}=&\bm{B}Z X_t+ \rho\cdot \bm{B}\bm{W}\bm{Z}X_{t-1}+...\\
&+ \rho^{t-1}\cdot\bm{B} (\prod_{\tau=1}^{t-1} \bm{W})  \bm{Z}X_{1}.
\end{aligned}
\end{equation}
Using the same way as above, we have
\begin{equation}
\begin{aligned}
&||f^{lin}_t(\bm{W},\bm{A})-f^{lin, \tau}_t(\bm{W},\bm{A})||_F\\
&\leq\sum_{k=\tau}\sum_{i+j=k-1}||\bm{B}||\cdot \rho^{k}\cdot ||\bm{W}_0^{i}||\cdot||\bm{W}_0^{j}|| \cdot||\bm{W}||\cdot||\bm{A}_0||\\
&+\sum_{k=\tau}||\bm{B}||\cdot \rho^{k}\cdot ||\bm{W}_0^{k}||\cdot ||\bm{A}||\\
&\leq \tau\rho_0^\tau\cdot8\frac{\sqrt{m}}{(1-\rho_0)^3}
\end{aligned}
\end{equation}
from Lemma \ref{lc}.

\QEDA

\section*{Existence: Proof of Theorem \ref{mkt}}
Firstly we briefly introduce the main steps of the proof.

1) From Lemma \ref{l9}, for all $t\in [T]$ and $\bm{W},  \bm{W}' \in B(\bm{W}_0, \omega)$, $\bm{A},  \bm{A}' \in B(\bm{A}_0, \omega)$ with $\omega \leq \omega_0$,
\begin{equation}
\begin{aligned}
&||f_t(\bm{W},\bm{A})-f_t(\bm{W}_0,\bm{A}_0)-\nabla_{W} f_t(\bm{W}_0, \bm{A}_0)\cdot [\bm{W}-\bm{W}_0]\\
& -\nabla_{A} f_t(\bm{W}_0, \bm{A}_0)\cdot [\bm{A}-\bm{A}_0] ||\leq 768\frac{\sqrt{m}\omega^2}{(1-\rho_0)^5}. \notag
\end{aligned}
\end{equation}
Therefore we consider the linearization function
\begin{equation}
\begin{aligned}
f_t^{lin}(\bm{W}, \bm{A})= &f_t(\bm{W}_0,\bm{A}_0)+\nabla_{W} f_t(\bm{W}_0, \bm{A}_0)\cdot [\bm{W}-\bm{W}_0]\\
& +\nabla_{A} f_t(\bm{W}_0, \bm{A}_0)\cdot [\bm{A}-\bm{A}_0]. \notag
\end{aligned}
\end{equation}
We have
\begin{equation}
\begin{aligned}
&||f_t(\bm{W},\bm{A})- f_t^{lin}(\bm{W}, \bm{A})||\leq  \cO(\frac{\sqrt{m}\omega^2}{(1-\rho_0)^5}).\notag
\end{aligned}
\end{equation}

2) From Lemma \ref{cut},
\begin{equation}
||f^{lin}_t(\bm{W},\bm{A})-f^{lin, T_{max}}_t(\bm{W},\bm{A})||\leq  \cO(\frac{\sqrt{m}\tau \rho_0^{\tau}}{(1-\rho_0)^3}). \notag
\end{equation}
Therefore we only need to consider
\begin{equation}
\begin{aligned}
&f^{lin, T_{max}}_t(\bm{W},\bm{A})=\sum _{t_0=0}^{T_{max}} \rho^{t_0}\bm{B} (\prod_{\tau=1}^{t_0} \bm{W}_0)  \bm{A}_0X_{t-t_0}\\
&+\sum_{t_0=0}^{T_{max}} \sum_{t_1+t_2=t_0}  \rho^{t_0}\bm{B}(\prod_{\tau=1}^{t_1-1}\bm{W}_0) [\bm{W}-\bm{W}_0]  (\prod_{\tau=1}^{t_2-1} \bm{W}_0)  \bm{A}_0X_{t-t_0}\\
&+\sum_{t_0=0}^{T_{max}}  \rho^{t_0}\bm{B}(\prod_{\tau=1}^{t_0-1}\bm{W}_0)[\bm{A}-\bm{A}_0]X_{t-t_0}. \notag
\end{aligned}
\end{equation}

3) Finally we set
\begin{equation}
\begin{aligned}
&\bm{W}^*-\bm{W}_0 \\
=&\sum_{t_m=1}^{T_{max}} (\rho^{-1})^{t_m-1}  \sum_{t_1+t_2=t_m-1}\{(\prod_{\tau=1}^{t_1-1}\bm{W}_0)\}^T\bm{B}^T \bm{P}^1_{t_1}\{\bm{G}\bm{C}^{t_m-1}\bm{D} \\
&-(\rho^2)^{t_m-1}/(t_m-1)\cdot \bm{B}\{\prod_{\tau=1}^{t_m} \bm{W}_0\}\bm{A}_0 \} \bm{P}^2_{t_2-1}\bm{A}^T_0 (\prod_{\tau=1}^{t_2} \bm{W}_0), \notag
\end{aligned}
\end{equation}
\begin{equation}
\bm{A}^*-\bm{A}_0=\{(\prod_{\tau=1}^{t_m-1}\bm{W}_0)\}^T\bm{B}^T \bm{P}^1_{t_m-1}[\bm{G}\bm{D}-\bm{B}\bm{A}_0], \notag
\end{equation}

where
\begin{equation}
\begin{aligned}
&\bm{P}^1_{t_1}= \{\bm{B}(\prod_{\tau=1}^{t_1-1}\bm{W}_0)(\prod_{\tau=1}^{t_1-1}\bm{W}_0)^T\bm{B}^T\}^{-1},\\
&\bm{P}^2_{t_2}= \{\bm{A}_0^T(\prod_{\tau=1}^{t_2-1}\bm{W}_0)^T(\prod_{\tau=1}^{t_2-1}\bm{W}_0)\bm{A}_0\}^{-1}.
\end{aligned}
\end{equation}

Then from Lemma \ref{lr}, \ref{lr2} and  \ref{lr3},
\begin{equation}\label{e55}
f^{lin, T_{max}}_{t}(\bm{W}^*,\bm{A}^*)\approx \sum _{t_0=0}^{T_{max}} \bm{G} (\prod_{\tau'=1}^{t_0} \bm{C})  \bm{D}X_{t-t_0}.
\end{equation}
With probability at least $1-exp(-\Omega(\log^2 m))$,
\begin{equation}
||f^{lin, T_{max}}_{t}(\bm{W}^*,\bm{A}^*))- \widetilde{y}_t|| \leq   \cO(\frac{ b\cdot d^2 c_\rho T^{3}_{max}\log m}{m^{1/2}}) +\frac{c_\rho \rho^{T_{max}}}{1-\rho}), \notag
\end{equation}
and we can show  $||\bm{W}^*-\bm{W}_0||_F, ||\bm{A}^*-\bm{A}_0||_F\leq \cO(b\cdot  T^2_{max}/\sqrt{m}) $ using Lemma \ref{lr2}. The theorem follows.

\begin{theo}\label{ex}
Consider a linear system:
\begin{equation}
\begin{aligned}
&p_{t}=\bm{C}p_{t-1}+\bm{D}x_t,\\
&\widetilde{y}_t=\bm{G}p_t.
\end{aligned}
\end{equation}
with $||\bm{C}^k\bm{D}||<c_\rho\cdot \rho^k$ for any $k\in \mathbb{N}$,  $\rho<1$, $\bm{D}\in \mathbb{R}^{d_p\times d}$, $p_t\in \mathbb{R}^{d_p}$,  $\bm{C}\in \mathbb{R}^{d_p\times d_p} $ $\bm{G}\in \mathbb{R}^{d_y\times d_p}$.

For given data $\{x_t, y_t\}$, and any $0<\rho<1$,  if $m> m^*$,  with probability at least $1-exp(-\Omega(\log^2 m))$, there exist $\bm{W}^*, \bm{A}^*$ satisfying that for all $t$,
\begin{equation}
\begin{aligned}
||f_{t}(\bm{W}^*,\bm{A}^*))- \widetilde{y}_t||_F &\leq \cO(\frac{\sqrt{m}(\rho_0)^{T_{max}}}{(1-\rho_0)^3})\\
&+ \cO(\frac{ b\cdot d^2 c_\rho T^{3}_{max}\log m}{m^{1/2}})  +\frac{c_\rho \rho^{T_{max}}}{1-\rho}),
\end{aligned}
\end{equation}
and
\begin{equation}
||\bm{W}^*-\bm{W}_0||_F, ||\bm{A}^*-\bm{A}_0||_F\leq 2c_\rho b\cdot T^2_{max}/\sqrt{m}.
\end{equation}
\end{theo}
{\bf Proof:}

Firstly, note that from Lemma \ref{l9}.
\begin{equation}
\begin{aligned}
&||f_t(\bm{W},\bm{A})-f_t(\bm{W}_0,\bm{A}_0)-\nabla_{W} f_t(\bm{W}_0, \bm{A}_0)\cdot [\bm{W}-\bm{W}_0]\\
& -\nabla_{A} f_t(\bm{W}_0, \bm{A}_0)\cdot [\bm{A}-\bm{A}_0] ||\\
&\leq \cO(\frac{\sqrt{m}\omega^2}{(1-\rho_0)^5}).
\end{aligned}
\end{equation}
Therefore let
\begin{equation}
\begin{aligned}
f_t^{lin}(\bm{W}, \bm{A})= &f_t(\bm{W}_0,\bm{A}_0)+\nabla_{W} f_t(\bm{W}_0, \bm{A}_0)\cdot [\bm{W}-\bm{W}_0]\\
& +\nabla_{A} f_t(\bm{W}_0, \bm{A}_0)\cdot [\bm{A}-\bm{A}_0].
\end{aligned}
\end{equation}
We have
$$||f_t(\bm{W}, \bm{A})-f_t^{lin}(\bm{W}, \bm{A}) ||\leq  \cO(\frac{\sqrt{m}\omega^2}{(1-\rho_0)^5}).$$
Note that
\begin{equation}
||\bm{W}^*-\bm{W}_0||_F, ||\bm{A}^*-\bm{A}_0||_F\leq \cO(b\cdot T^2_{max}/\sqrt{m}).
\end{equation}
We can show
$$||f_t(\bm{W}^*, \bm{A}^*)-f_t^{lin}(\bm{W}^*, \bm{A}^*) ||\leq  \cO(\epsilon/b).$$
Now, from Lemma \ref{cut},
\begin{equation}
||f^{lin}_t(\bm{W},\bm{A})-f^{lin, T_{max}}_t(\bm{W},\bm{A})||\leq \cO(\frac{\sqrt{m}\tau \rho_0^{\tau}}{(1-\rho_0)^3}).
\end{equation}
We only need to consider
\begin{align}
&f^{lin, T_{max}}_t(\bm{W},\bm{A})=\sum _{t_0=0}^{T_{max}} \rho^{t_0}\bm{B} (\prod_{\tau=1}^{t_0} \bm{W}_0)  \bm{A}_0X_{t-t_0}\notag \\
&+\sum_{t_0=0}^{T_{max}} \sum_{t_1+t_2=t_0}  \rho^{t_0}\bm{B}(\prod_{\tau=1}^{t_1-1}\bm{W}_0) [\bm{W}-\bm{W}_0] \notag \\
&\cdot  (\prod_{\tau=1}^{t_2-1} \bm{W}_0)  \bm{A}_0X_{t-t_0}\notag \\
&+\sum_{t_0=0}^{T_{max}}  \rho^{t_0}\bm{B}(\prod_{\tau=1}^{t_0-1}\bm{W}_0)[\bm{A}-\bm{A}_0]X_{t-t_0}. \notag
\end{align}
 Finally we set
\begin{align}
&\bm{W}^*-\bm{W}_0 \notag \\
&=\sum_{t_m=1}^{T_{max}} (\rho^{-1})^{t_m-1} \sum_{t_1+t_2=t_m-1}\{(\prod_{\tau=1}^{t_1-1}\bm{W}_0)\}^T\bm{B}^T \bm{P}^1_{t_1}\notag \\
&\cdot \{\bm{G}\bm{C}^{t_m-1}\bm{D} -(\rho^2)^{t_m-1}/(t_m-1)\cdot \bm{B}\{\prod_{\tau=1}^{t_m} \bm{W}_0\}\bm{A}_0 \}\notag \\
&\cdot  \bm{P}^2_{t_2-1}\bm{A}^T_0 (\prod_{\tau=1}^{t_2} \bm{W}_0), \notag
\end{align}

\begin{equation}
\bm{A}^*-\bm{A}_0=\{(\prod_{\tau=1}^{t_m-1}\bm{W}_0)\}^T\bm{B}^T \bm{P}^1_{t_m-1}[\bm{G}\bm{D}-\bm{B}\bm{A}_0],
\end{equation}
where
\begin{equation}
\begin{aligned}
&\bm{P}^1_{t_1}= \{\bm{B}\{(\prod_{\tau=1}^{t_1-1}\bm{W}_0)(\prod_{\tau=1}^{t_1-1}\bm{W}_0)^T\bm{B}^T\}^{-1},\\
&\bm{P}^2_{t_2}= \{\bm{A}_0^T(\prod_{\tau=1}^{t_2-1}\bm{W}_0)(\prod_{\tau=1}^{t_2-1}\bm{W}_0)\bm{A}_0\}^{-1}.
\end{aligned}
\end{equation}

Firstly we need to bound $||\bm{W}^*-\bm{W}_0||_F$ and $||\bm{A}^*-\bm{A}_0||_F$.

When $t_1, t_2\leq T_{max}$,  note that $\bm{P}^1_{t_1}$ and $\bm{P}^2_{t_2}$ are symmetric matrices. From Lemma \ref{lr}.
\begin{equation}
\begin{aligned}
&||\bm{P}^1_{t_1}||\leq \frac{2}{ m},\\
&||\bm{P}^2_{t_2}||\leq 2.
\end{aligned}
\end{equation}
Meanwhile $(\rho^{-1})^{t_m-1} ||\bm{G}\bm{C}^{t_m-1}\bm{D}|| \leq c_\rho $ since $\rho(\bm{C})\leq \rho$. Moreover, from Lemma \ref{lr2},
\begin{equation}
 ||\bm{B} (\prod_{\tau'=1}^{t_0} \bm{W})  \bm{A}||\leq  b,
 \end{equation}
 for all $t_0\leq T_{max}$.
Therefore
\begin{equation}
||\bm{W}^*-\bm{W}_0||_F, ||\bm{A}^*-\bm{A}_0||_F\leq 2c_\rho b\cdot T^2_{max}/\sqrt{m}.
\end{equation}

From lemma \ref{lr3}, when $t\neq t'$, $t, t'\leq \tau$,
\begin{equation}
\begin{aligned}
|(u_1)^T\bm{B}(\prod_{\tau=1}^{t}\bm{W}_0)\cdot& \{(\prod_{\tau=1}^{t'}\bm{W}_0)\}^T\bm{B}^Tv_1|\\
&\leq 24\tau d^2 \sqrt{m}\log m,
\end{aligned}
\end{equation}
and
\begin{equation}
|(u_2)^T\bm{A}^T_0(\prod_{\tau=1}^{t}\bm{W}_0)^T\cdot \{(\prod_{\tau=1}^{t'}\bm{W}_0)\}\bm{A}_0v_2|\leq 24\tau\frac{d^2\log m} {\sqrt{m}}.
\end{equation}
This shows

\begin{align}
&f^{lin, T_{max}}_t(\bm{W}^*,\bm{A}^*)\notag \\
=&\sum _{t_0=0}^{T_{max}} \rho^{t_0}\bm{B} (\prod_{\tau=1}^{t_0} \bm{W}_0)  \bm{A}_0X_{t-t_0}\notag \\
&+\sum_{t_0=0}^{T_{max}} \sum_{t_1+t_2=t_0}  \rho^{t_0}\bm{B}(\prod_{\tau=1}^{t_1-1}\bm{W}_0)\notag \\
&\cdot \sum_{t_m=1}^{T_{max}} (\rho^{-1})^{t_m-1} \sum_{t_1+t_2=t_m-1}\notag \\
&\cdot\{(\prod_{\tau=1}^{t_1-1}\bm{W}_0)\}^T\bm{B}^T \bm{P}^1_{t_1}\{\bm{G}\bm{C}^{t_m-1}\bm{D}\notag  \\
&-(\rho^2)^{t_m-1}/(t_m-1)\cdot \bm{B}\{\prod_{\tau=1}^{t_m} \bm{W}_0\}\bm{A}_0 \} \bm{P}^2_{t_2-1}\bm{A}^T_0 (\prod_{\tau=1}^{t_2} \bm{W}_0)\notag \\
&\cdot  (\prod_{\tau=1}^{t_2-1} \bm{W}_0)  \bm{A}_0X_{t-t_0}\notag \\
&+\sum_{t_0=0}^{T_{max}}  \rho^{t_0}\bm{B}(\prod_{\tau=1}^{t_0-1}\bm{W}_0)\{(\prod_{\tau=1}^{t_m-1}\bm{W}_0)\}^T\bm{B}^T\notag \\
& \cdot \bm{P}^1_{t_m-1}[\bm{G}\bm{D}-\bm{B}\bm{A}_0] X_{t-t_0}. \notag
\end{align}

Combining all the above results, Lemma \ref{lr2}  and
$$\widetilde{y_t}=\sum _{t_0=0}^{t-1} \bm{G} (\prod_{\tau=1}^{t_0} \bm{C})  \bm{D}X_{t-t_0} ,$$
we have
\begin{equation}
\begin{aligned}
&||f^{lin, T_{max}}_{t} (\bm{W}^*,\bm{A}^*)-\sum_{t_0=0 }^{T_{max}-1} \bm{G} (\prod_{\tau=1}^{t_0} \bm{C} )  \bm{D}X_{t-t_0}||\\
&\leq \cO(T_{max}^2\cdot 2b \cdot c_\rho \cdot T_{max}\frac{d^2\log m} {\sqrt{m}}),
\end{aligned}
\end{equation}
\begin{equation}
\begin{aligned}
||f^{lin, T_{max}}_{t}&(\bm{W}^*,\bm{A}^*))- \widetilde{y_t}|| \\
&\leq  \cO(\frac{ b\cdot d^2 c_\rho T^{3}_{max}\log m}{m^{1/2}}) +\frac{c_\rho \rho^{T_{max}}}{1-\rho}).
\end{aligned}
\end{equation}
The theorem follows.

\QEDA

Using Theorem \ref{ex}, from the definition of $T_{max}$, $$m>m^*,$$
and $\rho< \rho_0,$
\begin{equation}
T_{max}>\Theta(\frac{1}{\log(\frac{1}{\rho})} )\cdot \{ \log(\frac{1}{1-\rho}) +\log b) +\log(\frac{1}{\epsilon} ) \}),
\end{equation}
 we have
\begin{equation}
||f_{t}(\bm{W}^*,\bm{A}^*))-  \widetilde{y_t}|| \leq \cO(\frac{\epsilon}{l_0\cdot(1+2b)}) .
\end{equation}

Thus
$L(y_{t},f_{t}(\widetilde{\bm{W}}^*, \bm{A}^*))- L(y_{t}, \widetilde{y}_{t})\leq \cO(\epsilon).$ Theorem \ref{mkt} follows.

}
\end{document}